\documentclass[preprint,12pt]{elsarticle}

\usepackage{amssymb}
\usepackage{amsmath}
\usepackage[ruled,linesnumbered]{algorithm2e} 
\usepackage{hyperref} 
\hypersetup{colorlinks=true}
\usepackage{multirow} 
\usepackage{graphicx}
\usepackage{stfloats} % float figure
\usepackage{float} % float table

\usepackage{xcolor} 
\definecolor{darkblue}{RGB}{0, 0, 139} % RGB 值

\journal{Robotics and Autonomous Systems}

\begin{document}

\begin{frontmatter}


\title{Robotic Manipulation Framework Based on Semantic Keypoints for Packing Shoes of Different Sizes, Shapes, and Softness\tnoteref{label666}}
\tnotetext[label666]{This work was supported by the National Youth Science Foundation of China under Grant No. 52205017. The Aviation Foundation under Grant No.2022Z050052001.}

\author[label1]{Yi Dong\corref{co-first authors}} %% Author name
\ead{dongyi@nuaa.edu.cn}
\author[label1,label2]{Yangjun Liu\corref{co-first authors}} %% Author name
\ead{marcoliu@nuaa.edu.cn}
\author[label1]{Jinjun Duan\corref{corresponding}} %% Author name
\ead{duan-jinjun@nuaa.edu.cn}
\author[label1]{Yang Li} %% Author name
\ead{yangli@nuaa.edu.cn}
\author[label1]{Zhendong Dai} %% Author name
\ead{zddai@nuaa.edu.cn}

\cortext[co-first authors]{The authors contributed equally to the work.}
\cortext[corresponding]{Corresponding author.}

\affiliation[label1]{organization={College of Mechanical and Electrical Engineering, Nanjing University of Aeronautics and Astronautics},%Department and Organization
            %addressline={}, 
            city={Nanjing},
            postcode={210016}, 
            %state={},
            country={China}}

\affiliation[label2]{organization={State Key Laboratory of Internet of Things for Smart City and Department of Electromechanical Engineering, University of Macau},%Department and Organization
            %addressline={}, 
            city={Macau},
            postcode={999078}, 
            %state={},
            country={China}}

%% Abstract
\begin{abstract}
With the rapid development of the warehousing and logistics industries, the packing of goods has gradually attracted the attention of academia and industry.
The packing of footwear products is a typical representative paired-item packing task involving irregular shapes and deformable objects.
Although studies on shoe packing have been conducted, different initial states due to the irregular shapes of shoes and standard packing placement poses have not been considered. 
This study proposes a robotic manipulation framework, including a perception module, reorientation planners, and a packing planner, that can complete the packing of pairs of shoes in any initial state. 
First, to adapt to the large intraclass variations due to the state, shape, and deformation of the shoe, we propose a vision module based on semantic keypoints, which can also infer more information such as size, state, pose, and manipulation points by combining geometric features.
Subsequently, we not only proposed primitive-based reorientation methods for different states of a single deformable shoe but also proposed a fast reorientation method for the top state using box edge contact and gravity, which further improved the efficiency of reorientation. 
Finally, based on the perception module and reorientation methods, we propose a task planner for shoe pair packing in any initial state to provide an optimal packing strategy.
Real-world experiments were conducted to verify the robustness of the reorientation methods and the effectiveness of the packing strategy for various types of shoes.
In this study, we highlight the potential of semantic keypoint representation methods, introduce new perspectives on the reorientation of 3D deformable objects and multi-object manipulation, and provide a reference for paired object packing.
\end{abstract}

%% Keywords

\end{frontmatter}


\section{Introduction}
\label{introduction}

\begin{figure}[htbp]
  \centering
  \includegraphics[scale=0.4]{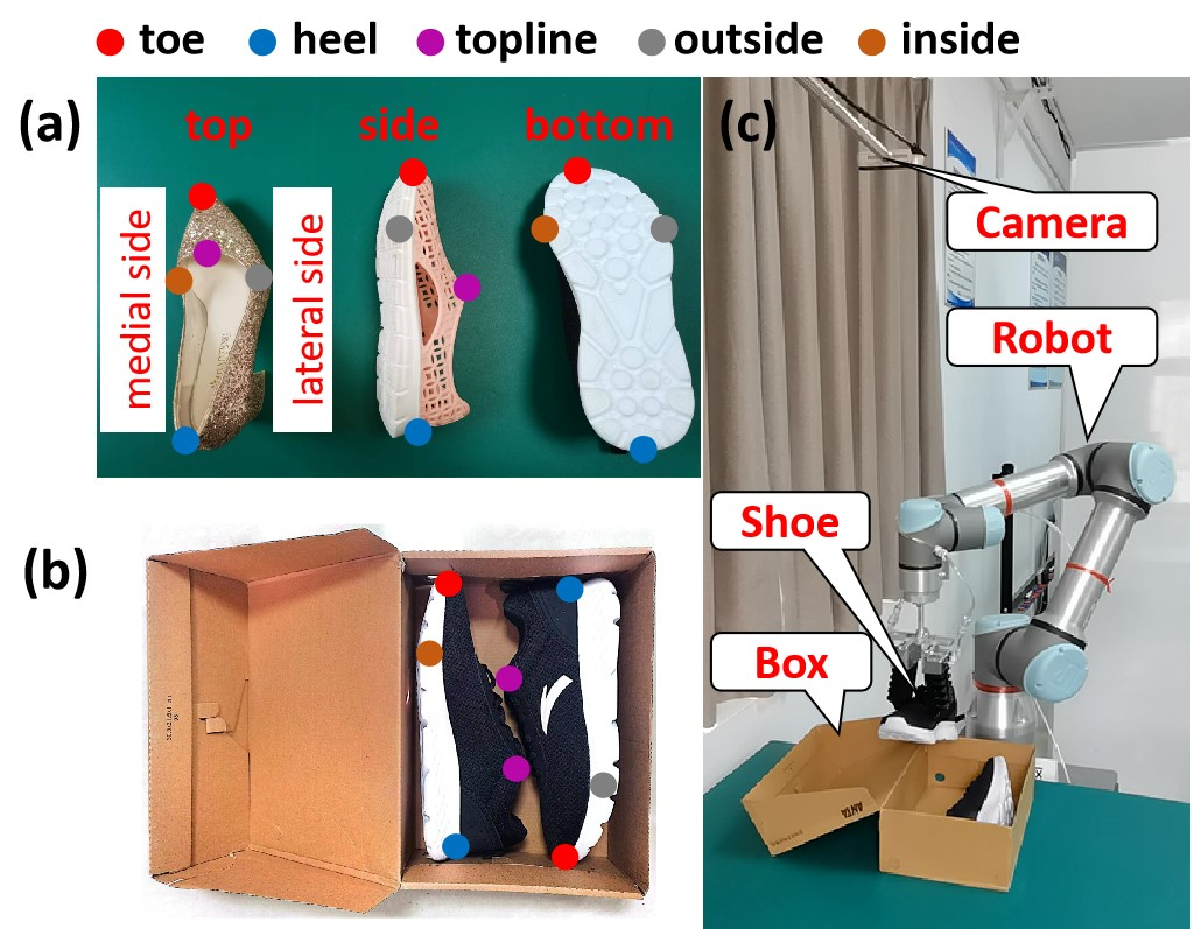}
  \caption{Initial and target states of shoe packing and the corresponding robotic system. (a) Three initial states: top, side, and bottom, with the distribution of five keypoints: toe, heel, topline, outside (lateral side), and inside (medial side). (b) Target pose of a pair of shoes in the box. (c) Robotic shoe packing system.}
  \label{intro}
\end{figure}

Packing and placing items into containers is a pre-step in the logistics transportation stage.
The packing stage involves objects of different shapes and softness, which is challenging during the packing process.
Related studies exist on this topic, such as those on the packing of rigid irregular objects \cite{huang2022planning} \cite{wang2021dense} and flexible objects \cite{bahety2023bag} \cite{ma2022action}.
However, the main goal of these packing tasks is efficient space utilization, and objects are assumed to be directly graspable and placeable.
Some special objects, such as shoes shown in Fig. \ref{intro} (a), exhibit different stable poses (referred to as states), and some states cannot reach the target pose (Fig. \ref{intro} (b)) by one-time grasping and placing. 
Reorientation is necessary, and more stringently, the relative pose between objects (such as shoes and boxes, and between shoes) must be considered (Fig. \ref{intro} (b)). 
However, current shoe packing research \cite{perez2018automation}, \cite{gracia2017robotic} does not consider arbitrary initial states or standard target poses of shoes.
Additionally, there is a large intracategory shape variation, and the shoe may deform during manipulation, which also leads to perception and manipulation challenges.

For the shoe-packing task, the first thing we need to consider is the representation of an object.
The shape differences between the different states and types of shoes are significant, and the deformation is uncertain. 
The information required in the packing process far exceeds the grasping pose information, which is a problem that must be solved via the perception component.
Common object representation methods include 6D pose \cite{lin2023vi} \cite{lin2024instance} \cite{zou2024learning} \cite{periyasamy2023yolopose}, dense representations \cite{zhang2024vision} \cite{amir2021deep} \cite{liu2019gift}, and keypoint representations \cite{manuelli2019kpam} \cite{gao2021kpamsc} \cite{gao2021kpam}. 
Among them, keypoint representation has unique advantages: In addition to being able to adapt to large intracategory variations, it can provide more details to facilitate the manipulation of objects and is a low-dimensional representation \cite{manuelli2019kpam} \cite{gao2021kpamsc} \cite{gao2021kpam} \cite{fang2024keypoint}.
Current keypoint-based manipulation tasks use keypoints to characterize key features of objects \cite{deng2024general} \cite{sundaresan2023kite} \cite{luo2022skp} \cite{robson2022keypoint} \cite{gao2021kpamsc} \cite{wang2020learning} \cite{manuelli2019kpam} or relationships \cite{huang2024rekep} \cite{liu2024moka} \cite{xu2021affordance} \cite{qin2020keto} between objects to directly guide manipulation, without further mining the information contained in the keypoints. 
In fact, by combining keypoints and geometric features of objects within the class, we can not only calculate the grasping pose but also infer the size, state, and manipulation points of the shoe, which are all necessary information for packing.
In this study, we propose a keypoint-based perception module that can obtain shoe states and poses and can be used to manipulate primitives for shoe reorientation.

The packing task must also consider the placement of the objects. 
In this task, the placement poses are clearly specified rather than arbitrary. 
In actual grasping and placement, the shoes are in different states. Under the constraints of the movement space and environment (e.g., desktop and container) of the robot, one pick-and-place method may not be able to achieve the specified pose \cite{xu2023learning} \cite{mishra2024reorientdiff} \cite{wada2022reorientbot} \cite{xu2022efficient}. 
At this point, reorientation is crucial. Common reorientation methods rely on primitives, such as pushes and topples, all of which use the desktop \cite{shome2019towards} \cite{wan2015reorientating} \cite{chen2023visual} \cite{chen2022system}. 
However, these studies are all conducted on rigid objects, whereas shoes have flexible uppers that cause certain difficulties in reorientation. 
In this study, we propose primitive-based reorientation methods for shoes in different states, considering the deformation of the shoes.
To improve the efficiency and robustness of shoe packing, we additionally propose a reorientation method using box edge constraints and gravity for the top state. Furthermore, we examine the effect of the placement order on the success rate.

In this study, we complete the shoe packing task by placing a pair of shoes in any initial state in a shoe box according to standard packing requirements. 
The manipulation framework comprises a keypoint-based perception module, reorientation planners, and a packing task planner. 
To enhance the robustness and generalizability of the framework, these modules consider the shape, state diversity, and deformability of the shoes.
Compared to existing shoe packing methods \cite{perez2018automation}, \cite{gracia2017robotic}, \cite{qin2020keto}, \cite{wang2020learning}, our main contributions are as follows:

\begin{itemize}

\item 
A vision module that combines keypoints and geometric features can provide more information. 
The proposed visual perception module can not only adapt to large intracategory variations, but also obtain more valuable information by combining geometric information, such as size, state, pose, and manipulation point information that facilitates subsequent manipulation.

\item 
A primitive-based reorientation approach for 3D deformable objects using the compliance of a soft gripper. 
According to different states of the shoe, the proposed primitive uses the interference between the soft gripper and the shoe to complete the reorientation.

\item 
An additional reorientation method that exploits the contact between objects provides a new perspective on multi-object manipulation.
In the packing task, we propose an efficient reorientation method for the top state that exploits the contact with box edges and gravity.

\item 
A complete robotic manipulation framework for the packing of paired objects. 
Based on any initial state combination, the framework offers an optimal state transition strategy and a high-success-rate placement sequence to efficiently complete packing.

\end{itemize}

The remainder of this paper is organized as follows. 
Related studies are summarized in Section \ref{Related Work}. 
The problem is described in Section \ref{Problem Statement}. 
Section \ref{Proposed Solution} provides our solution for the shoe-packing task, including visual perception (Section \ref{Visual Perception Approach}), shoe reorientation methods (Section \ref{Keypoint-based Shoe Manipualation}), and a task planner for the entire task (Section \ref{Keypoint-based Task Planner}).
Section \ref{Experiments and Results} presents several experiments on the advancement of the perception pipeline, robustness of shoe reorientation methods, and feasibility of the packing strategy. 
Finally, Sections \ref{Discussion} and \ref{Conclusion} discuss and conclude the paper, respectively.

\section{Related work}
\label{Related Work}

\subsection{Robotic packing}

Prior to logistics transportation, goods are typically packaged in two stages \cite{vieira2021packing} \cite{gracia2017robotic} \cite{morales2014bimanual} \cite{balatti2021flexible}. 
The first stage involves packing individual items into containers (e.g., boxes), whereas the second stage involves packing these smaller boxes into larger containers for easy transportation.
Compared with the second stage \cite{tresca2022automating} \cite{agarwal2020jampacker} \cite{yang2023heuristics} \cite{jia2022robot} \cite{zhao2022learning}, the first stage remains an open area of research because it involves goods with varying shapes and rigidities. 
For rigid objects, the focus has shifted from packing cubic objects \cite{zhou2022method} \cite{wang2020robot} \cite{shome2019tight} \cite{shome2019towards} to packing irregularly shaped items \cite{mojtahedi2024experimental} \cite{huang2022planning} \cite{wang2021dense} \cite{wang2019stable}. 
In the case of flexible objects, research has addressed the packing of long elastic items \cite{ma2022action} and accumulated fabric objects \cite{bahety2023bag}, with packing solutions extended to flexible bags \cite{bahety2023bag} \cite{chen2023autobag}. 
However, these methods do not apply to shoe-packing tasks because shoes are rigid-soft coupled objects with multiple states, and both reorientation and pairing poses should be considered during the packing process.

Research closely related to shoe packing has led to the development of shoe packing pipelines, which include object detection, object grasping, trajectory planning with collision avoidance, and operator interactions using force/torque sensors \cite{perez2018automation} \cite{gracia2017robotic} \cite{morales2014bimanual}. 
These studies have pioneered the packing of footwear products. 
However, they did not account for different initial states and final relative poses of the shoes in the shoe box. 
Although our previous study \cite{dong2023robotic} considered the target relative poses of shoes, the initial state of the shoes was assumed as the top state as opposed to a random state. 
Additionally, the success rate of the reorientation method, which relies solely on contact and gravity, is relatively low for certain types of shoes.

\subsubsection{Vision modules in robotic packing}
% \textbf{Vision Module: }
Recent advances in robotic packing systems rely heavily on robust visual perception. For rigid objects, many works estimate 6D pose using RGB-D fusion techniques \cite{wang2021dense} \cite{shome2019towards}, which typically assume relatively stable geometry across instances. For deformable objects, recent studies instead detect structural keypoints \cite{ai2024robopack}, grasping points \cite{bahety2023bag} \cite{chen2023autobag}, or contours \cite{bahety2023bag} using deep networks or foundation models to handle large shape variations.
Unlike prior work that uses visual modules solely for isolated tasks such as pose estimation or grasp prediction, our method leverages semantic keypoints to extract richer geometric and task-relevant information—such as object state, size, and reorientation point. 
These visual cues are specifically tailored to the requirements of shoe-packing, enabling downstream planners to reason over deformability, initial states, and pairing constraints in a unified visual-manipulation pipeline.

% foucus the first stage of packing objects
\subsubsection{Planning strategies for robotic packing}
% \textbf{Packing Planner: }
Packing planners aim to optimize object placement for space efficiency and stability. 
Existing methods range from rule-based heuristics and combinatorial algorithms \cite{wang2021dense} \cite{wang2020robot} \cite{wang2019stable} to learning-based approaches \cite{huang2022planning} that predict placements from object and scene features. 
Beyond passive arrangement, recent planners actively reshape object configurations using manipulation primitives—such as compact placement \cite{zhou2022method}, toppling \cite{shome2019towards} \cite{shome2019tight}, and regrasping \cite{wan2019regrasp} —to ensure feasible, tight, and task-aware packing.
However, most prior work assumes single, rigid objects and overlooks the challenges posed by deformable, state-dependent, 3D paired items like shoes.
These methods rarely integrate reorientation methods or multi-object reasoning. 
In contrast, our planner incorporates semantic keypoints and reorientation methods to enable flexible and adaptive strategies for packing shoe pairs with varying shapes, softness, and initial configurations.

\subsection{Object representations for manipulation}

In robotics, object-representation methods generally include pose, dense, and keypoint representations \cite{xue2023useek} \cite{manuelli2019kpam} \cite{gao2021kpam}. 
Specifically, 6D pose estimation provides a global representation of objects, making it useful for tasks such as object grasping. However, it struggles with large intracategory shape variations because it assumes similar geometric structures within the same category, limiting its adaptability to deformable objects \cite{lin2023vi} \cite{lin2024instance} \cite{zou2024learning}.
Dense Visual Descriptors capture fine-grained information from every pixel, offering robustness to lighting and viewpoint changes; however, their high computational cost and challenges in handling significant object deformations limit their practical use \cite{zhang2024vision} \cite{amir2021deep} \cite{liu2019gift}. 
Among these methods, keypoint representations can adapt to large intraclass variations and provide richer local details than pose representations, therefore providing a low-dimensional alternative to dense representations.

Owing to the unique advantages mentioned above, keypoint representation has been applied to various manipulation tasks, ranging from simple pick-and-place tasks \cite{manuelli2019kpam} \cite{gao2021kpamsc} to door opening \cite{wang2020learning}, tool manipulation \cite{qin2020keto} \cite{xu2021affordance} \cite{robson2022keypoint}, and the handling of flexible objects \cite{huo2022keypoint}. 
Structural keypoints can only serve as references for action planning, whereas semantic keypoints carry semantic information that can guide higher-level task planning. 
This enables a more generalized and abstract representation of object manipulation through low-dimensional encoding.
Given these advantages, manipulation tasks based on imitation learning \cite{fang2024keypoint} \cite{riou2023temporal} \cite{xue2023useek} and large models \cite{huang2024rekep} \cite{li2024skt} \cite{deng2024general} \cite{sundaresan2023kite}, which have emerged in recent years, also use semantic keypoint representation methods to quickly learn and adapt to new tasks. 
In these studies, keypoints were mainly used to guide manipulation directly. However, the visual information conducted by the keypoints was not further explored.
To adapt to the large intraclass variants and deformable characteristics of shoes, in this study, we also used a semantic keypoint representation.
The difference is that, in this study, we further extracted the information carried by key points, such as size, state, posture, and reorientation action point, by combining the geometric features within the class for packing tasks.

\subsection{Object reorientation}

Reorientation can be achieved using two main approaches: in-hand manipulation and extrinsic manipulation. 
In-hand manipulation uses a multi-fingered hand in free space with control strategies to reorient an object \cite{chen2023visual} \cite{mack2023soft} \cite{chen2022system} \cite{chen2021simple} \cite{da2017stable}.
However, extrinsic manipulation simplifies the reorientation process using environmental constraints, thereby reducing the complexity of the gripper and control strategies \cite{dafle2014extrinsic} \cite{xu2023learning} \cite{wan2015reorientating}.
The earliest reorientation research can be traced back to the 1980s \cite{tournassoud1987regrasping}, and it was implemented by exploiting desktops and manipulation primitives, including pushing \cite{lynch1996stable}, squeezing \cite{brost1985planning}, pivoting \cite{aiyama1993pivoting}, and toppling \cite{lynch1999toppling}. 
These studies mainly focused on stable poses and achieving reorientation using primitives. 
Subsequent research has used these primitives in the orientation state network graph \cite{farooqi1996reorientation} \cite{omata1997reorientation} and in the reorientation planning of objects for specific tasks \cite{wan2015reorientating} \cite{nguyen2016preparatory} \cite{wan2019regrasp} \cite{mishra2024reorientdiff}. 
Recently, many studies used learning-based methods to extend stable pose reorientation to the arbitrary pose reorientation of objects with arbitrary shapes \cite{hou2018fast} \cite{xu2022efficient} \cite{wada2022reorientbot}.
To the best of our knowledge, all these methods assume the object being manipulated to be rigid. 
Although reorienting tasks exist for flexible materials \cite{kristek2012orienting}, they mainly focus on 2D flexible materials, such as paper, and cannot be applied to 3D deformable objects, such as shoes in this task. 
In this study, we propose reorientation methods for 3D irregular deformable objects in different states. 

In addition to desktops, when other constraints such as containers appear in the environment, they can be used to reorient objects. 
For example, a recent study \cite{xu2023learning} examined the use of containers in an environment to regrasp and reorient objects. 
During the experimental test, the object was placed in the container in advance to expose the graspable part and then grasped to achieve reorientation, such as flipping. 
Based on \cite{xu2023learning}, the box in this task also creates external support for the reorientation of shoes. However, the difference is that we propose a new strategy that uses the contact of the box edge and gravity to achieve rapid reorientation by a one-time grasp and placement. 
Furthermore, in multi-object manipulation tasks such as packing, contact between objects exists naturally instead of being provided additionally.

\section{Problem and notation}
\label{Problem Statement}

We address the scenario of a manipulator tasked with packing a pair of shoes using an RGBD camera, as shown in Fig. \ref{intro} (c). 
The challenge with shoe packing is that shoes display different sizes, shapes, and softnesses, thereby leading to difficulties in perception and manipulation. 
To place a pair of shoes in any initial state into the box as required, designing a robotic manipulation framework that can accurately identify the key features and poses of the shoes, enable robust shoe reorientation, and guide the packing process is necessary.

To address this issue, we decompose the shoe-packing framework into three distinct parts.

\textbf{Visual Perception:} 
First, the semantic keypoints of the shoes, denoted by $K_{shoe}$ = $\{P_{toe}$, $P_{heel}$, $P_{topline}$, $P_{outside}$, $P_{inside}\}$ where \( P_i \in \mathbb{R}^3 \), are extracted from an RGBD image using a deep neural network. 
Based on these keypoints and the structure of the footwear product, the state of the shoe $S_{shoe}$ (which can be top, side, or bottom) and its pose $P_{shoe}$ are inferred; this provides the necessary information for subsequent manipulation tasks.

\textbf{Shoe Reorientation:} 
Once the shoe state is determined, a reorientation method must be designed to change the shoe state to a state convenient for packing. 
This includes primitive-based and contact-based methods, where the manipulation primitive can be represented as a tuple ($P_1$, $P_2$) with \( P_i \in \text{SE}(3) \), which depicts the starting and ending poses of the primitive, respectively. 

\textbf{Packing Planner:} 
The last step is the packing task planning strategy $\pi_{task}$, which integrates the aforementioned visual perception information and reorientation methods, can complete the state recognition and conversion of pairs of shoes, and complete the packing of shoes in an optimal order to ensure the efficiency and accuracy of the packing process.

\section{Proposed solution}
\label{Proposed Solution}

\subsection{Visual perception approach}
\label{Visual Perception Approach}

\begin{figure}[htbp]
  \centering
  \includegraphics[width=13cm]{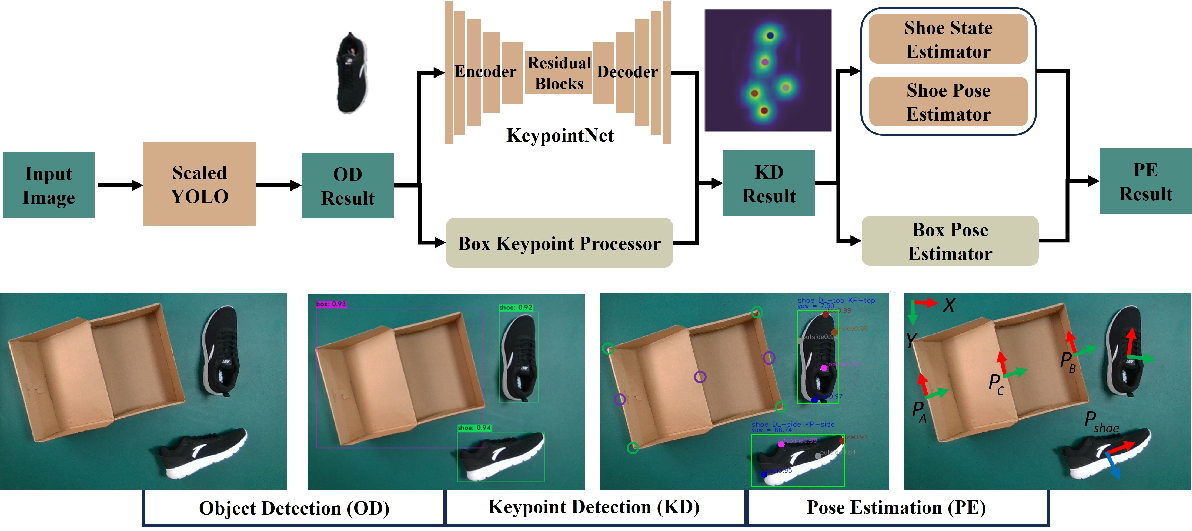} %
  \caption{Overview of the visual perception module including object detection, keypoint detection, and pose estimation. In the first phase, YOLO obtains the classes and bounding boxes of the object. Next, a novel-designed KeypointNet model with an image processor is used to detect keypoints of the shoes and the box. Finally, based on these detected keypoints, the states and poses of the shoe and box are calculated for the shoe packing process. The code for the perception module is publicly available at \href{https://gitlab.com/NUAADongyi/keypoint-based-shoe-packing}{https://gitlab.com/NUAADongyi/keypoint-based-shoe-packing}.}
  \label{visual_pipeline}
\end{figure}

To adapt to large intraclass variations and extract more visual information, we propose a keypoint-based visual perception module. 
Fig. \ref{visual_pipeline} shows the pipeline for detecting categories, keypoints, states, and poses of objects in multi-object manipulation tasks. 
This pipeline comprises three main parts: object detection, keypoint detection, and pose estimation.

\subsubsection{Object detection}

YOLO \cite{Wang_2021_CVPR} was applied to predict the locations of horizontal bounding boxes with target object categories based on the confidence scores. 
We trained a YOLO network that could precisely detect bounding boxes. 
Given that the two downstream modules capture images cropped by the generated bounding boxes as inputs, accurate bounding boxes pave the way for satisfactory performance in downstream tasks.
The model was trained on a multi-object dataset comprising 1,172 images (912 for training and 260 for testing) from three categories: box, paper, and shoe. 
The objects were randomly placed on tabletops with diverse patterns and photographed from multiple viewpoints, and each image was annotated with horizontal bounding boxes and category labels.

\subsubsection{Shoe keypoint detection}

We propose a novel keypoint detection network called KeypointNet, which takes an RGB image of a shoe and outputs several pixel-wise heat maps of the keypoints. 
In the shoe keypoint detection task, five heat maps of semantic keypoints on the toe, heel, inside, outside, and topline of the shoe (Fig. \ref{intro}) were generated, irrespective of the structure, appearance, and pose.

\textbf{Model Architecture: }
KeypointNet has an encoder-decoder-like architecture. 
The encoder consisted of five convolutional layers while the decoder comprised five transposed convolutional layers. 
Each layer was subjected to batch normalization. 
To extract better features and avoid the degradation of the deep neural network \cite{he2016deep}, five stacked basic residual blocks were inserted between the encoder and decoder. 
We systematically designed the kernel size, stride, and padding of each layer such that the input images and output heat maps exhibited the same height and width. 
At the end of the decoder, a dropout layer was added before the output layer to counter the tendency towards co-adaptions of feature channels and overfitting of deep networks \cite{srivastava2014dropout}.

\textbf{Loss Function: }
Considering the precision of the predicted keypoint locations and the distribution of confidence scores on heatmaps, we employed an overall loss $\mathcal{L}_{all}$, which was a linear combination of the average normalized 2D Euclidean distance of keypoints $\mathcal{L}_{ned}$ and the mean squared error of the corresponding heatmaps $\mathcal{L}_{mse}$. 
$\mathcal{L}_{all}$ is defined as
\begin{equation}\label{eq1}
    \mathcal{L}_{all} = \alpha \mathcal{L}_{mse} + (1-\alpha)\mathcal{L}_{ned},
\end{equation}
where $\alpha$ denotes a hyper-parameter of the overall loss. 
This dual-loss framework leverages both dense heatmap supervision $\mathcal{L}_{mse}$ and sparse keypoint regression $\mathcal{L}_{ned}$. 
The latter accelerates convergence by directly penalizing localization errors, which $\mathcal{L}_{mse}$ alone may not optimally rectify due to its pixel-averaged nature.
The training process exhibited an evident increase in the convergence rate when $\alpha$ was assigned a value of 0.618. 
$\mathcal{L}_{mse}$ and $\mathcal{L}_{ned}$ are expressed as (\ref{eq2}) and (\ref{eq3}), respectively.
\begin{equation}\label{eq2}
    \mathcal{L}_{mse} = \frac{1}{N_{mse}H_{img}W_{img}}\sum\limits_{kij}(S_{kij} - \hat{S}_{kij})^2,
\end{equation}
\begin{equation}\label{eq3}
    \mathcal{L}_{ned} = \frac{1}{N_{ned}}\sum\limits_{k \in \mathbb{K}}\sqrt{(x_k - \hat{x}_k)^2 + (y_k - \hat{y}_k)^2},
\end{equation}
where $S_{kij}$ denotes the ground-truth confidence score at location ($i$, $j$) of the $k$th heatmap and $\hat{S}_{kij}$ denotes the corresponding predicted score. 
The normalized location with the highest confidence score on each heatmap was considered as the keypoint, denoted as ($x_k$, $y_k$), and the corresponding predicted location was denoted as ($\hat{x}_k$, $\hat{y}_k$). 
${N_{mse}}$, $N_{ned}$, $H_{img}$, and $W_{img}$ constitute the number of heatmaps, keypoints, height, and width of an image-heatmap pair, respectively. 
In addition, we consider only keypoints whose ground-truth confidence scores are labeled 1 when calculating $\mathcal{L}_{ned}$ as follows:
\begin{equation}\label{eq4}
    \mathbb{K} = \{k|\max{S_{kij}}=1\}.
\end{equation}

Hence, unseen keypoints in the current pose are not considered when computing $\mathcal{L}_{ned}$.

\textbf{Implementation Details: }
The KeypointNet model was trained on a dataset containing 557 single-shoe images (367 for training and 190 for testing), with shoes captured under diverse backgrounds, styles, and poses, and annotated with manually labeled keypoints. 
To improve robustness, we applied data augmentation techniques including random padding, flipping, rotation, background replacement, and color filtering \cite{khalifa2022comprehensive}.
The network was trained with a batch size of eight on a remote NVIDIA Tesla V100 GPU for 340 epochs (approximately 17 h) using the PaddlePaddle framework \cite{ma2019paddlepaddle} \cite{bi2022paddlepaddle}.  
The Adam optimizer was used with an initial learning rate of 1e-3, which decreased to 1e-4 at the 290th epoch. 
When the trained model was deployed on an NVIDIA GeForce RTX 3080 GPU, an inference speed of approximately 36 fps was obtained.

\subsubsection{State and pose estimation}

Based on the above keypoints, the state and pose of the object can be further deduced according to geometric features.

\textbf{Shoe State Classification: }
To place a pair of shoes into a box with the proper relative pose, the initial state of the shoes should be considered to select an appropriate reorientation method.
Generally, because of the irregular shape of the shoe, its initial state is finite, including the top, side, and bottom (Fig. \ref{intro} (a)).
The three states can be obtained through postprocessing according to the predicted keypoints and can be distinguished by the following rules (Equation \ref{s1}).

\begin{equation}\label{s1}
\begin{aligned}
S_{shoe} =
    \begin{cases}
    top, & \forall P_i (P_i \in K_{shoe}) \neq 0 \\
    side, & P_{inside}=0 || P_{outside}=0 \\
    bottom, & P_{topline}=0 
    \end{cases}
\end{aligned}
\end{equation}

% \subsubsection{Shoe Pose Estimation}
\textbf{Shoe Pose and Grasp Pose Estimation: } 
The shoe pose $P_{shoe}$ and corresponding grasp pose $G_{shoe}$ can be calculated based on the detected semantic keypoints. 
The coordinate system of the shoe is illustrated in Fig. \ref{visual_pipeline}. The X-axis direction is aligned from the heel point to the toe point, and the Z-axis direction is perpendicular to the sole of the shoe. 
The position of the shoe is determined by the midpoint between $P_{toe}$ and $P_{heel}$. 
The roll angle of the shoe is determined by its state, and its yaw angle, which can be calculated using Equation \ref{eq5}, is used as the grasping angle $\theta_{grasp}$.
The grasping position of the shoe $P_{grasp}$ is set to the position of the shoe.
Thus, the grasping pose of the shoe is defined as $G_{shoe} = (P_{grasp}, \theta_{grasp})$.
\begin{equation}\label{eq5}
    \theta_{grasp} = arctan(\overrightarrow{H_0T_0} \times \overrightarrow{HT}, \overrightarrow{H_0T_0} \cdot \overrightarrow{HT})
\end{equation}

We assumed that the initial X-axis direction of the shoe $\overrightarrow{H_0T_0}$ was aligned with that of the camera. 
$\overrightarrow{HT}$ is the $ x $ axis direction of the current coordinate system of the shoe.

For shoe state classification and pose estimation, we trained deep-learning models, including ResNet \cite{he2016deep} and CenterPose \cite{lin2022icra:centerpose}, to serve as comparative baselines, which will be discussed in Section \ref{Results on multi-object dataset}. 
These models were trained on two additional datasets that share the same 557 single-shoe images (367 for training and 190 for testing) as the keypoint dataset. 
Each image was labeled with states and pose parameters (yaw and roll angles).

\textbf{Box Pose Estimation: }
The corners of the boxes can be obtained based on their structural features. 
However, the box with a lid has more than four corners from different perspectives (Fig. \ref{visual_pipeline}).
Based on our previous work \cite{dong2023robotic}, both the minimum external rectangle and the convex hull are necessary to calculate the keypoints.
In the convex hull, the four right-angle corners can be approximately represented as the contour points closest to the four corners of the minimum bounding rectangle of the box.
Based on the four keypoints and the geometric features of the box, the midpoint and slope on the three parallel sides can be calculated to determine the box position and orientation.
The box is represented by $P_{box} = \{P_A, P_B, P_C\}$ and $(P_i \in {SE}^3)$ as shown in Fig. \ref{visual_pipeline}.

\subsection{Shoe reorientation}
\label{Keypoint-based Shoe Manipualation}

\subsubsection{Shoe toppling} 

To achieve the target configuration, side + side state with inside + outside keypoint pairing (Fig. \ref{intro} (b), a single grasp-and-place operation is often insufficient and a toppling primitive becomes necessary. 
First, since the target state for packing is the side state, shoes in the top or bottom state must be toppled accordingly. 
Second, even when both shoes are already in the side state, their keypoint pairing may still be inside + inside or outside + outside, which does not meet the requirement. Thus, a toppling primitive is also required for the side state. 
In summary, the toppling algorithm must be applicable to shoes in all three states: top, side, and bottom.
Given different initial states, the objective is to calculate the start and end poses of toppling ($P_1, P_2$), which can change the state of the shoe.

\begin{figure}[ht]
  \centering
  \includegraphics[scale=0.50]{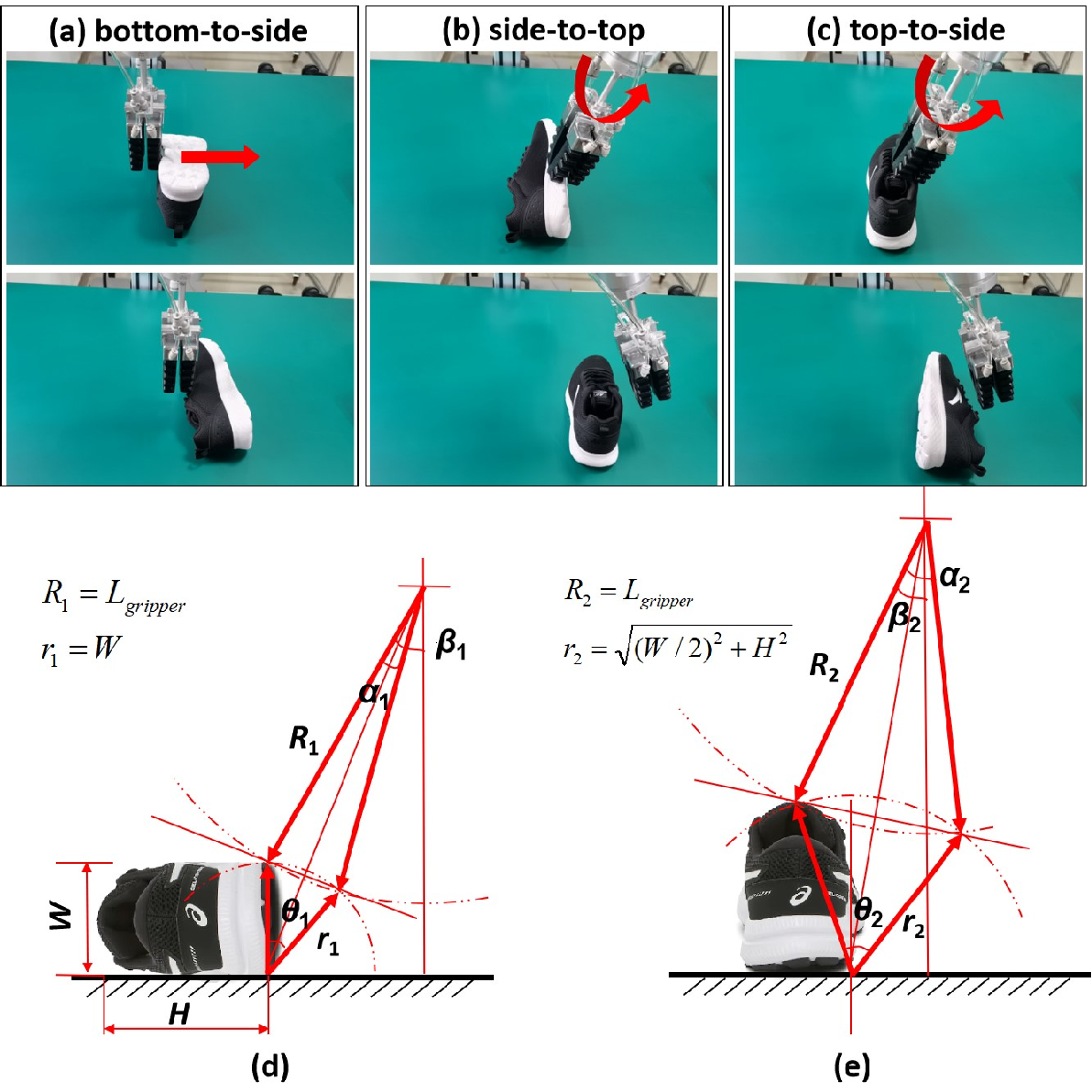}
  \caption{Toppling primitives for reorienting the shoe. (a) Push action to change the state from the bottom to the side; (b) and (c) show the rotation action used to change the state from side to top and their reverse; and (d) and (e) show the kinematic models of (b) and (c), respectively.}
  \label{shoe_toppling}
\end{figure}

According to previous studies \cite{shome2019tight} \cite{shome2019towards} \cite{lynch1999toppling}, toppling involves two phases: rolling and settling.
% referencez: Toppling_manipulation
In the rolling phase, the robot pushes the object to rotate around the toppling edge until the projection of its center of mass on the table exceeds the toppling edge.
Simultaneously, the angle of rotation is called the toppling angle.
During the settling phase, an object falls to a new position under the force of gravity.
In this shoe-packing task, the toppling edge is parallel to the X-axis of the shoe according to its shape. 
In the bottom state, the center of mass is above the toppling edge. Therefore, the shoe topples under a small disturbance from the outside, such as a push action (Fig. \ref{shoe_toppling} (a)).
However, when the shoe is in the side or top state, it may slide or deform (soft uppers) when pushed because of its unique structure and material.
As shown in Fig. \ref{shoe_toppling} (b) and (c), if the robot uses the rotation action to topple the shoe with the interference in these cases, its success rate significantly increases.

The kinematic models of toppling for the side and top states are shown in Fig. \ref{shoe_toppling} (d) and (e), respectively.
The toppling pose ($P_1, P_2$) corresponds to the initial angle $\beta_1$ and rotation angle $\alpha_1$ of the gripper for the side state, and $\beta_2$ and $\alpha_2$ for the top state.
These angles can be calculated from the width $W$ and height $H$ of the shoe and the respective toppling angles $\theta_1$ and $\theta_2$.
The detailed process is presented in Algorithm \ref{algorithm1}.

Algorithm \ref{algorithm1} presents the toppling algorithm for different initial shoe states.
It returns the toppling poses ($P_1, P_2$) that can change the shoe state.
When the state is at the bottom, the toppling poses can be calculated using the keypoints of the shoe, including the inside point $P_{inside}$ and outside point $P_{outside}$.
When the shoe is in the side state (Fig. \ref{shoe_toppling} (b)), the toppling direction must be determined first.
Subsequently, according to the keypoints of the shoe $K_{shoe}$ and the height of the table $H_{table}$, the width $W$ and height $H$ of the shoe can be obtained and used to determine the toppling angle $\theta_1$.
Based on the kinematic models shown in Fig. \ref{shoe_toppling}, the initial angle $\beta_1$ and the rotation angle $\alpha_1$ of the gripper can be obtained using Equations \ref{eq9} and \ref{eq10}. 
To avoid collisions, the maximum initial angle $\beta_{1max}$ calculated from Equation \ref{eq10s} is used to verify the feasibility of $\beta_1$.
Similar to the side-state transformation, the initial angle $\beta_2$ (where $\beta_2 < \beta_{2max}$, the maximum allowable initial angle) and the rotation angle $\alpha_2$ for the top state can also be derived using Equations \ref{eq11} - \ref{eq12s}.
These angles are then used to compute the toppling pose ($P_1, P_2$).

\begin{algorithm}
  \caption{Toppling algorithm}\label{algorithm1}
  \SetKwData{Bottom}{bottom}\SetKwData{Side}{side}
  
  \SetKwFunction{Push}{Push}
  \SetKwFunction{topplingDirection}{toppleDirection}
  \SetKwFunction{sideSize}{sideSize}
  \SetKwFunction{sideAngle}{sideAngle}
  \SetKwFunction{sidePose}{sidePose}
  \SetKwFunction{topSize}{topSize}
  \SetKwFunction{topAngle}{topAngle}
  \SetKwFunction{topPose}{topPose}
  
  \KwIn{shoe keypoints $K_{shoe}$, shoe pose $P_{shoe}$, shoe state $S_{shoe}$, table height $H_{table}, $gripper length $L_{gripper}$}
  \KwResult{start pose and end pose for toppling $(P_{1}, P_{2})$}

  \uIf(\tcp*[h]{$P_{inside},P_{outside} \in K_{shoe}$}){$S_{shoe} = \Bottom$}{
  $(P_{1}, P_{2})\leftarrow$ \Push{$P_{inside},P_{outside},P_{shoe}$}\;
  }
  \uElseIf(\tcp*[h]{$P_{top} \in K_{shoe}$}){$S_{shoe} = \Side$}{
  d $\leftarrow$ \topplingDirection{$P_{top}, P_{shoe}$}\;
  $(W, H)\leftarrow$ \sideSize{$K_{shoe},P_{shoe},H_{table}$}\;
  $\theta_1 \leftarrow \arctan(H/W)$\;
  $(\alpha_1, \beta_1)\leftarrow$ \sideAngle{$W, \theta_1, L_{gripper}$}\;
  $(P_{1},P_{2})\leftarrow$ \sidePose{$d,\alpha_1,\beta_1,K_{shoe},P_{shoe}$}\;
  }
  \Else{
  $(W, H)\leftarrow$ \topSize{$K_{shoe},P_{shoe},H_{table}$}\;
  $\theta_2 \leftarrow \arctan(W/H)$\;
  $(\alpha_2, \beta_2)\leftarrow$ \topAngle{$W, H, \theta_2, L_{gripper}$}\;
  $(P_{1}, P_{2})\leftarrow$ \topPose{$\alpha_2, \beta_2, K_{shoe}, P_{shoe}$}\;
  }
  
  \Return{$(P_{1}, P_{2})$} 
\end{algorithm}

\begin{equation}\label{eq9}
    W\sin\frac{\theta_1}{2}  = L_{gripper}\sin\frac{\alpha_1}{2}
\end{equation}
\begin{equation}\label{eq10}
    \beta_1 = \frac{\alpha_1}{2} + \frac{\theta_1}{2}
\end{equation}
\begin{equation}\label{eq10s}
    L_{gripper}\cos\beta_{1max} + W = L_{gripper}
\end{equation}
\begin{equation}\label{eq11}
    \sqrt{(\frac{W}{2})^2 + H^2} \sin\frac{\theta_2}{2}  = L_{gripper}\sin\frac{\alpha_2}{2}
\end{equation}
\begin{equation}\label{eq12}
    \beta_2 = \frac{\alpha_2}{2} + \frac{\theta_2}{2} - \arctan\frac{W/2}{H}
\end{equation}
\begin{equation}\label{eq12s}
    L_{gripper}\cos\beta_{2max} + H = L_{gripper}
\end{equation}

Soft grippers play an important role in primitive-based reorientation methods. 
To counteract shoe deformation, the end-effector must interfere with the shoe to successfully change its pose.
The compliance of the soft gripper can adapt well to interference and protect the shoe from damage.
However, when the upper part is very soft (especially the top state), these primitive-based methods may fail. Therefore, we also propose a contact-based reorientation method, which can further improve the efficiency and robustness of packing.

\subsubsection{Reorienting through contact} 

Inspired by the use of containers for object regrasping \cite{xu2023learning}, we propose a new method for reorientation through contact with a box edge, which can be completed with a single pick-and-place. 
In this packing task, the box provides a step with a height difference between the edge and bottom that can be used to reorient the shoe. 
Based on the target shoe pose shown in Fig. \ref{intro} (b), this method is suitable for the shoe with a top state.
This method is specifically designed for the top state and is not applicable to side or bottom states, as these either already satisfy the target condition or result in non-target orientations after reorientation.

\begin{figure}[htbp]
  \centering
  \includegraphics[scale=0.6]{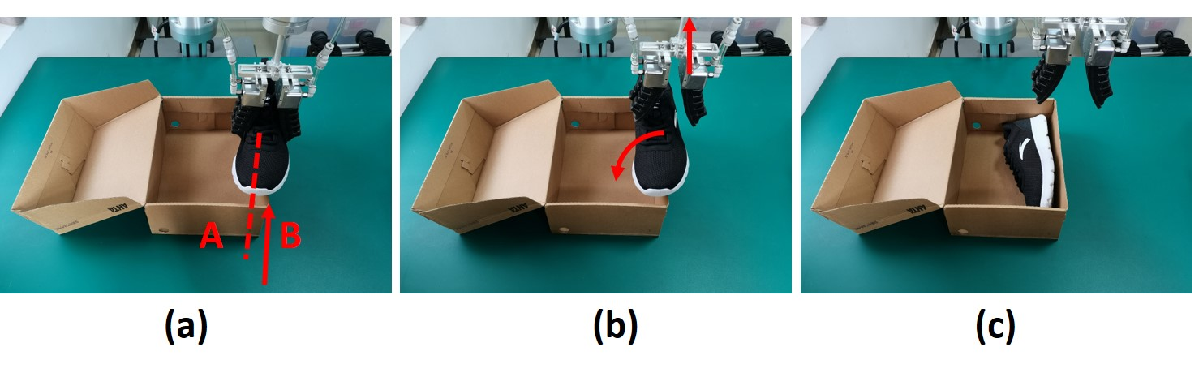}
  \caption{Reorienting the shoe through contact with the box. (a) Place the shoe on the edge of the box with an offset between the center line of the shoe and the edge. (b) The shoe rotates when the robot opens the gripper and moves up. (c) The final pose of the shoe after falling under gravity. }
  \label{placement}
\end{figure}

The reorientation process is similar to \cite{zhang1998automatic}, who proposed a reorientation method for 3D convex polyhedral parts using a sequence of step devices on a conveyor belt. 
The difference is that we focus on the reorientation of shoes with irregular shapes and softness in a single step (naturally formed by the edge of the box).
As shown in Fig. \ref{placement}, the robot initially places the shoe at point $P_B$ on the edge of the box and makes the midline of the shoe exhibit a certain offset from the edge of the box (Fig. \ref{placement} (a)). 
When the robot opens the gripper and moves up, the shoe rotates and slides, given that the moment is caused by gravity (Fig. \ref{placement} (b)). 
Finally, it collides with the bottom of the box and reaches a stable position (Fig. \ref{placement} (c)). 
The state of the shoe varies at the end of the process. 
Notably, the top state may go to either the side or the bottom state. 
Based on the quantitative analysis of previous studies \cite{zhang1998automatic}, for known shoes and boxes, the final state of the shoe can be controlled by regulating the amount of the offset. 
In Section \ref{Experiments on reorienting through a step}, we present the effects of the offset amount and placement order on the reorientation success rate for different types of shoes.

To clarify the roles of the two reorientation methods within the packing workflow, we explicitly define their functions in the two-stage packing process, which consists of a pre-placement stage and a placement stage. 
The primitive-based reorientation method operates during the pre-placement stage, where it adjusts the shoe’s pose into a state suitable for placement. 
In contrast, the contact-based reorientation method performs reorientation during the placement stage itself, effectively combining both reorientation and placement in a single action. 
While the latter is more efficient, it can only be applied to shoes initially in the top state. 
Therefore, in the following task planner module, we discuss how to intelligently combine both reorientation methods to achieve optimal packing efficiency. 

\subsection{Packing task planner} 
\label{Keypoint-based Task Planner}

\begin{figure}[htbp]
  \centering
  \includegraphics[scale=0.55]{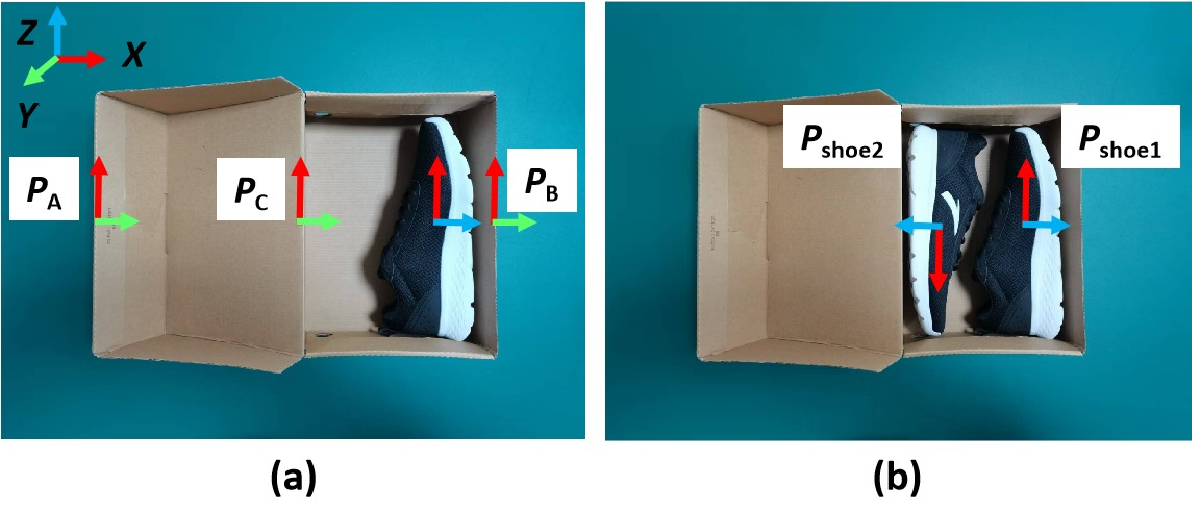}
  \caption{Placement poses during the packing of a shoe pair. (a) Pose of the first shoe with respect to the box. (b) Pose of the second shoe with respect to the first shoe.}
  \label{placement_pose}
\end{figure}

The aforementioned reorientation methods are used for a single shoe. However, in the shoe packing process, the state combination of a pair of shoes must be considered.
The final state of a pair of shoes is illustrated in Fig. \ref{intro} (b), and its symbolic representation is given by Equations \ref{eq6}-\ref{eq7}. 
The ultimate goal of the shoe packing task is to ensure that all shoes are placed in the final side + side configuration, with their keypoints aligned in an inside + outside manner. 
In addition, the relative orientation between the shoes and the box, as well as between the two shoes, must be carefully maintained (Fig. \ref{placement_pose}). 
As indicated in Equations \ref{eq8}-\ref{eq9}, the Z-axis of Shoe 1 $P_{shoe1}^Z$ should align with the Y-axis of the designated keypoint on the box $P_{B}^Y$, while the X-axis of Shoe 1 $P_{shoe1}^X$ should be opposite to that of Shoe 2 $P_{shoe2}^X$.

\begin{equation}\label{eq6}
    S_{shoe1}=side \wedge S_{shoe2}=side 
\end{equation}
\begin{equation}\label{eq7}
    P_{inside}=1 \wedge  P_{outside}=1
\end{equation}
\begin{equation}\label{eq8}
    P_{shoe1}^Z = P_{A}^Y
\end{equation}
\begin{equation}\label{eq9}
    P_{shoe1}^X = -P_{shoe2}^X
\end{equation}

\begin{figure}[ht]
  \centering
  \includegraphics[scale=0.55]{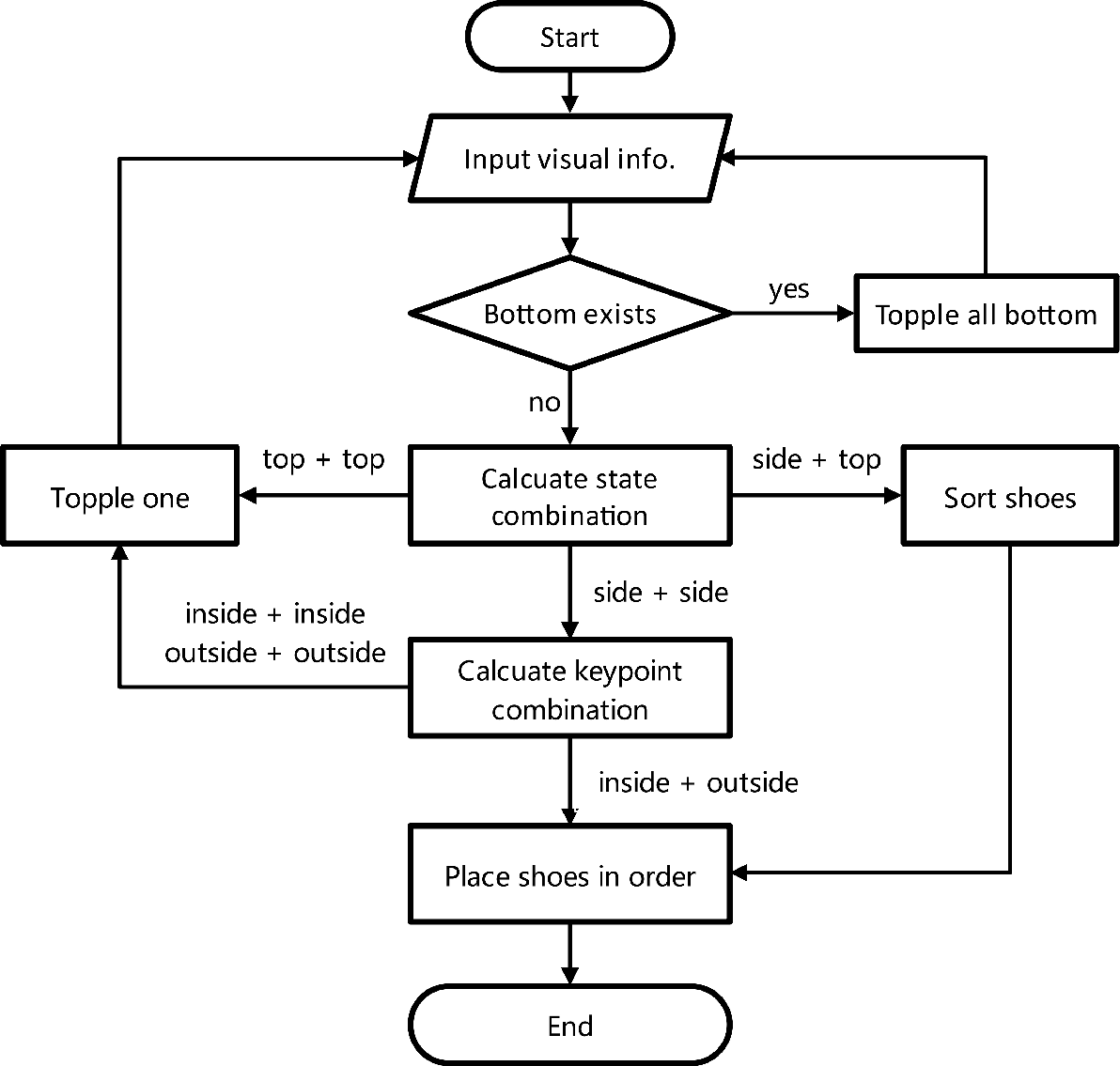}
  \caption{Flowchart of the task planning process for packing a pair of shoes based on different combinations of their initial states. The process is implemented using a state machine that determines whether reorientation is required and selects the appropriate packing strategy accordingly. }
  \label{state_combination}
\end{figure}

In this section, we formulate efficient packing strategies for different combinations of shoe pairs (Fig. \ref{state_combination}). 
For the packaging of a pair of shoes, it is necessary to consider both its state combination and keypoint combination. %, and even the placement order.
A pair of shoes may exhibit various state combinations, including top + top, top + side, top + bottom, side + side, side + bottom, and bottom + bottom.
Regardless of the initial combination, the final placeable states must be either in the side state or in the top state, with the latter being enabled for placement through the contact-based reorientation method.
Therefore, we first detect all shoes in the bottom state and convert them into the side state. 
After this reorientation step, the valid state combinations for a pair of shoes include side + side, side + top, and top + top.
When the side + top appears, packaging can be performed directly. The side can be grasped and placed directly, reorienting the top through contact with the box, and the subsequent experiments also further studied the impact of its placement order on the success rate.
The placement poses of the two shoes are illustrated in Fig. \ref{placement_pose} for reference.
For the side + side combination, if its keypoint combination satisfies Equation \ref{eq7}, the shoes can be grasped and placed directly.
Conversely, if its keypoint combination is two inside keypoints or two outside keypoints, then the state of one of the shoes must be changed.
In this case, applying a toppling primitive to either shoe will convert the combination into side + top, which is directly placeable.
For the top + top combination, two theoretical packaging methods exist.
The first method is to place both shoes by exploiting their contact with the box.
The second method is to change the state of one of the shoes, resulting in a side + top combination that can be placed directly.
As demonstrated in Section \ref{Experiments on reorienting through a step}, the second method yields a higher success rate and is therefore adopted in our framework.

\section{Experiments and results}
\label{Experiments and Results}

\subsection{Experimental setup}
\label{Experimental setup}

As mentioned in the previous section, the compliance of the soft gripper adapts to the interference with the deformable part of the shoe during the reorientation process and also to the irregular shape of the shoe during the grasping process. 
The robotic shoe packing system consists of a UR5e robot arm, a four-finger soft gripper (Rochu GC-4FMA6V5/ LS1–SMP2L–FCMR03), and an RGBD camera (Inter RealSense D415i) mounted above the robot arm (Fig. \ref{intro} (c)). 
When performing the experiments, the proposed solution was run on a desktop computer with an Intel i7 2.9 GHz CPU and an NVIDIA GeForce RTX 3080 GPU, which was sufficient for real-time keypoint estimation and robot control.
MoveIt was used for robot motion planning to precompute the motion before task implementation.
The ROS framework was used to integrate the perception module, planning module, and the robotic hardware system.
The code for the manipulation framework is publicly available at \href{https://gitlab.com/NUAADongyi/keypoint-based-shoe-packing}{https://gitlab.com/NUAADongyi/keypoint-based-shoe-packing}. 

\subsection{Evaluation of visual perception approach}

\subsubsection{Metrics for visual perception approach}

Four metrics were utilized to evaluate the performance of the visual perception algorithms on our multi-object dataset. 
The first was the metric for object detection. We adapted the average precision (AP) of Intersection over Union (IoU) with a threshold of 75\%, as described in \cite{Wang_2021_CVPR}.
To evaluate the position error of the detected shoe keypoints, we calculated the mean dimensionless Euclidean distance between the detected and ground-truth keypoints and used it as the second metric. 
Specifically, the Euclidean distance of each labeled and predicted keypoint pair was nondimensionalized by the ground-truth distance between the toe and heel keypoints of each shoe, thereby uniformly assessing the keypoint position error of shoes of different sizes. 
The precision of the shoe state classification was the third metric. 
We recognized a state prediction as a true-positive sample only when it was identical to its corresponding ground-truth label. 
For shoe pose recognition and estimation, we proposed the AP of yaw and roll with an error threshold as the fourth metric instead of reporting the AP of azimuth and elevation with a threshold, as in \cite{ahmadyan2021objectron} and \cite{lin2022icra:centerpose}. 
This is because the former directly represents the performance of object orientation estimation.

\subsubsection{Results on multi-object dataset}
\label{Results on multi-object dataset}

Our object detection model achieved high average precision at the 75\% IoU threshold for each category, i.e., 0.9995, 0.9805, and 0.9571 for box, paper, and shoe, respectively. 
In addition, the mean dimensionless position error of shoe keypoints predicted using KeypointNet was approximately 0.01774. 
Consequently, pose estimation benefitted from the precise object and keypoint detection results. 
The performance of the two aforementioned shoe-state classification methods is shown in Table \ref{state_classification}. 
Although the precision of the deep learning method in the top and side states exceeds that of the post-processing method based on KeypointNet, the latter clearly outperforms the former by approximately 6.67\% when the bottoms of the shoes face up. 
Moreover, keypoint-based methods can provide more information about shoes and can be used to explain reorientation methods and packing strategies.

\begin{table}[htbp]
\caption{Precision Comparison of Shoe State Classification Methods}
\label{state_classification}
\begin{center}
\resizebox{0.75\textwidth}{!}{ 

\resizebox{\linewidth}{!}{
\begin{tabular}{|c|c|c|c|c|}
\hline
\multirow{2}{*}{\textbf{Method}} & \multicolumn{4}{c|}{\textbf{Classification Precision}}\\
\cline{2-5}
& Top & Side & Bottom & Weighted Average\\
\hline
ResNet-34    &      1.0000           & 1.0000          & 0.9333          & 0.9867 \\
%Keypoint-based & 0.9432          & 0.9211          & \textbf{1.0000} & 0.9457\\ % model 0.016625
KeypointNet & \textbf{0.9565}  & \textbf{0.9722} & \textbf{1.0000} & \textbf{0.9715}\\ % model 0.011527
Difference &    -4.35\%           & -2.78\%         & 6.67\%          & -1.52\%\\ 
\hline
\end{tabular}
}
}
\end{center}
\end{table}

\begin{figure}[htbp]
  \centering
  %\framebox{\parbox{3in}{We suggest that you use a text box to insert a graphic (which is ideally a 300 dpi TIFF or EPS file, with all fonts embedded) because, in an document, this method is somewhat more stable than directly inserting a picture.}}
  \includegraphics[scale=0.34]{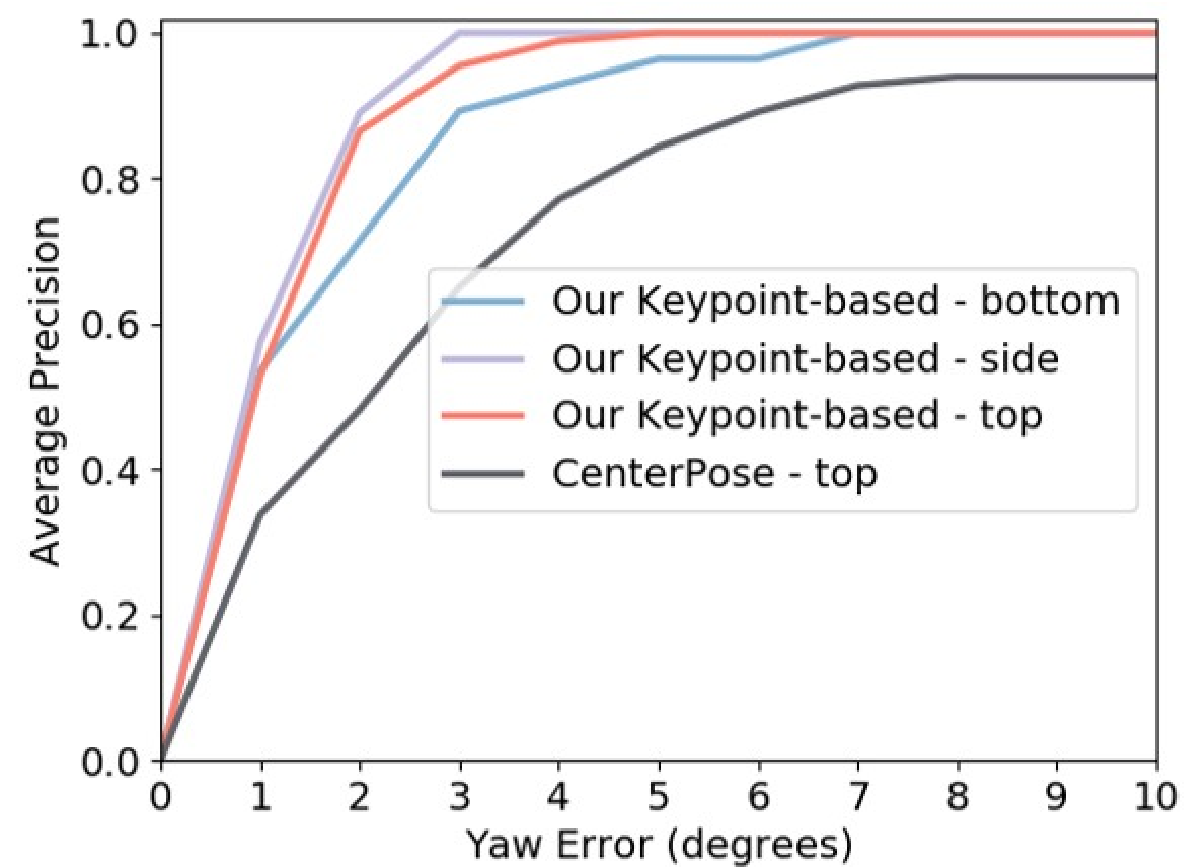}
  \caption{A comparison of average precision at a range of yaw errors.}
  \label{AP_Yaw}
\end{figure}

\begin{table}[htbp]
\caption{Orientation Estimation Comparison on the Shoe Testing Set}
\label{orientation_estimation}
\begin{center}
\resizebox{0.75\textwidth}{!}{

\resizebox{\linewidth}{!}{
\begin{tabular}{|c|c|c|c|}
\hline
\multirow{2}{*}{\textbf{Method}}
& \multirow{2}{*}{\begin{tabular}[c]{@{}c@{}} \textbf{Applicable for Shoes}\\\textbf{in Side/Bottom} \end{tabular}} 
& \multicolumn{2}{c|}{\textbf{Top state}}\\
\cline{3-4}
& & \begin{tabular}[c]{@{}c@{}} \textbf{Average} \\\textbf{Yaw Error} \end{tabular}
& \begin{tabular}[c]{@{}c@{}} \textbf{Average} \\\textbf{Roll Error} \end{tabular}\\
\hline
CenterPose                            & No  & 6.2766°          & Failed \\
KeypointNet                           & Yes & \textbf{1.0921°} & \textbf{4.2632°} \\
\hline
\end{tabular}
}
}
\end{center}
\end{table}

\begin{figure}[htbp]
  \centering
  \includegraphics[scale=0.45]{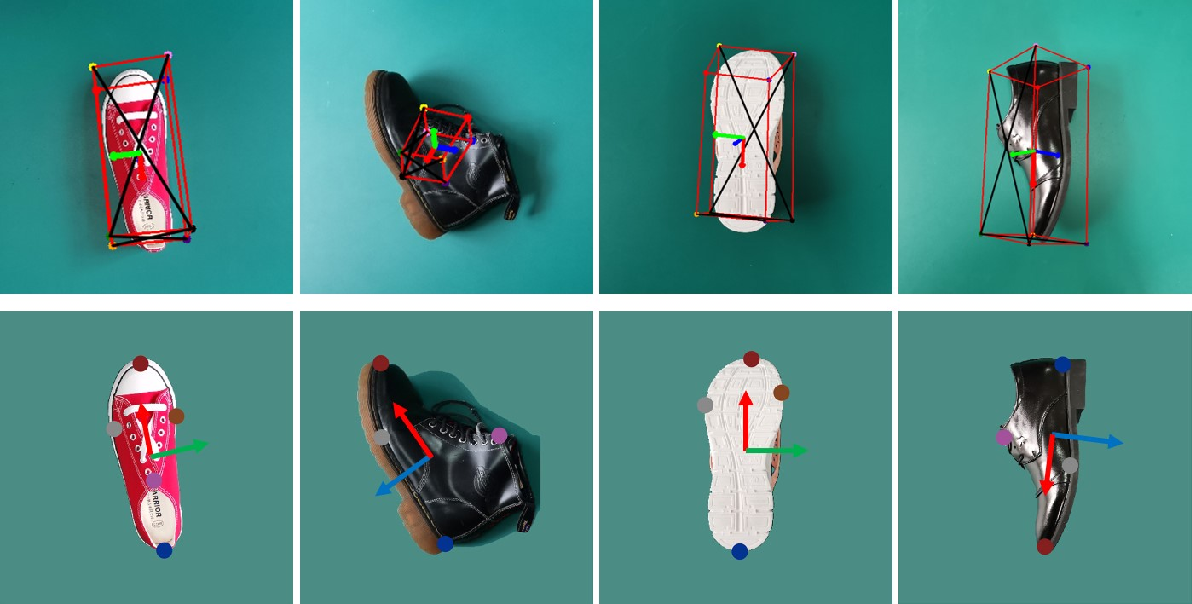}
  \caption{Qualitative pose estimation results of CenterPose (1st row) and our keypoint-based method (2nd row).}
  \label{qualitative_results}
\end{figure}

\begin{table}[htbp]
\caption{Inference Speed (FPS) of Different Methods for State Classification (SC) and Orientation Estimation (OE)}
\label{infer_speed}
\begin{center}
\resizebox{0.75\textwidth}{!}{
\resizebox{\linewidth}{!}{
\begin{tabular}{|c|c|c|c|c|}
\hline
\textbf{Method} & SC on 1050M & SC on 3080 & OE on 1050M & OE on 3080 \\
\hline
ResNet-34       & 66  & 80 & -- & -- \\
CenterPose      & -- & -- & 13  & 27 \\
KeypointNet  & \textbf{117} & \textbf{201} & \textbf{210} & \textbf{346}\\
\hline
\end{tabular}
}
}
\end{center}
\end{table}

The AP at the yaw and roll error thresholds is a comprehensive metric because shoe pose estimation relies on the results of the three upstream tasks. 
We compared our method with a state-of-the-art pose-estimation method, CenterPose. %\cite{lin2022icra:centerpose}. 
It was adapted to estimate the orientations of our shoe testing set (190 shoe images) for a fair comparison. 
Fig. \ref{AP_Yaw} demonstrates that our method outperforms the previous one for any threshold of yaw error of AP metric. 
As presented in Table \ref{orientation_estimation} and Fig. \ref{qualitative_results}, the CenterPose method usually fails to estimate the yaw and roll for shoes in the side and bottom states.
Moreover, the inference speed of our keypoint-based methods significantly exceeds that of the deep learning methods in the state classification and orientation estimation tasks as listed in Table \ref{infer_speed}.
For CenterPose, we only considered the time consumed in the postprocessing step, as opposed to the entire process described in \cite{lin2022icra:centerpose}, for an equitable comparison with our keypoint-based postprocessing method. 
We tested the algorithms on an NVIDIA GeForce GTX 1050 Mobile GPU (1050M) and an NVIDIA GeForce RTX 3080 GPU (3080), and obtained consistent results. 
The results indicate that our keypoint-based methods can achieve high inference speeds even on a low-performance GPU.

\subsection{Shoe reorientation experiments}

\begin{figure}[ht]
  \centering
  \includegraphics[scale=0.5]{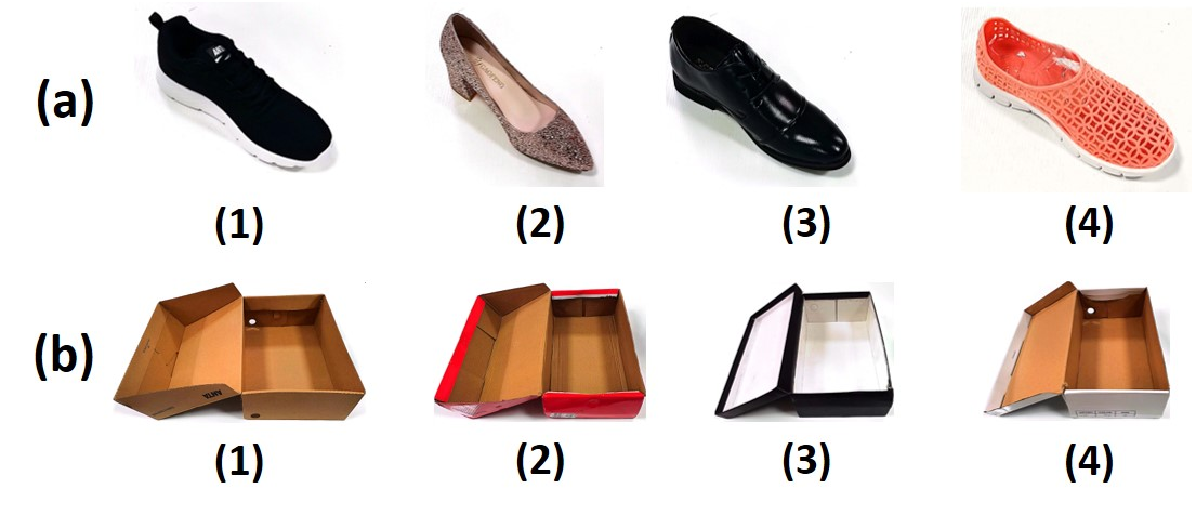}
  \caption{Different types of shoes and their matching boxes for testing. (a) The tested shoes include (1) sports shoes (281 mm long, 245 g per shoe), (2) high-heeled shoes (255 mm long, 255 g per shoe), (3) leather shoes (290 mm long, 398 g per shoe), and (4) sandal shoes (266.5 mm long, 235 g per shoe). (b) shows the matching boxes with different colors and sizes: (1) 300 mm $\times$ 220 mm $\times$ 110 mm, (2) 300 mm $\times$ 180 mm $\times$ 90 mm, (3) 330 mm $\times$ 205 mm $\times$ 115 mm, and (4) 315 mm $\times$ 190mm $\times$ 110 mm.}
  \label{shoe_types}
\end{figure}

To prove the robustness of our methods, the success rates of different reorientation methods were tested.
During the experiment, the tested shoes (Fig. \ref{shoe_types} (a)) include sports, high-heeled, leather, and sandal shoes, with their matching boxes shown in Fig. \ref{shoe_types} (b), respectively. 
These shoes exhibited large intracategory variations with different structures and materials.

\subsubsection{Shoe toppling experiments}

In the shoe toppling experiments, each state transformation for the different types of shoes was tested 10 times. 
As listed in Table \ref{toppling_results}, high-heeled and leather shoes do not have bottom states due to their unique structures.
One failure observed in sports shoe toppling was caused by inaccurate state estimation. 
The elastic upper part of sandal shoes can lead to over-rotation from the bottom to the side and then to the top state. Thus, it fails twice.

\begin{table}[htbp]
\caption{Experimental results on shoe toppling}
\label{toppling_results}
\begin{center}
\resizebox{0.75\textwidth}{!}{

\resizebox{\linewidth}{!}{
\begin{tabular}{|c|c|c|c|c|}
\hline
\textbf{\begin{tabular}[c]{@{}c@{}}Shoe\\ type\end{tabular}} & 
\textbf{\begin{tabular}[c]{@{}c@{}}Sports\\ shoes\end{tabular}} & 
\textbf{\begin{tabular}[c]{@{}c@{}}High-heeled\\ shoes\end{tabular}} & 
\textbf{\begin{tabular}[c]{@{}c@{}}Leather\\ shoes\end{tabular}} & 
\textbf{\begin{tabular}[c]{@{}c@{}}Sandal\\ shoes\end{tabular}} \\ \hline
\begin{tabular}[c]{@{}c@{}}bottom2side\end{tabular}    & 9/10  & N/A  & N/A   & 8/10 \\
side2top                                               & 9/10  & 1/10 & 10/10 & 10/10 \\
top2side                                               & 10/10 & 9/10 & 7/10  & 10/10 \\ \hline
\end{tabular}
}
}
\end{center}
\end{table}

Regarding the side state, failures in sports shoes were caused by inaccurate keypoint detection and exhibited a success rate of 9/10. An inaccurate keypoint detection can result in incorrect toppling positions or directions.
High-heeled shoes have very low success rates of approximately 1/10 because
of over-rotation during toppling.
Owing to the large support surface of high-heeled shoes in the side state, a large moment is required for toppling. However, a large moment for toppling is difficult to control.

Concerning the top state, similar to most failed cases, inaccurate keypoint detection led to the failures of high-heeled and leather shoes. Owing to the reflectivity of the upper of leather shoes, the success rate was approximately 7/10.

Generally, all failures are caused by errors in the perception or over-rotation of the shoes. 
Ultimately, the success rate of the toppling primitive is determined by the shoe structure and material. 
Notably, the structure of high heels determines the stability of the side state, and the reflectiveness of the leather shoe material poses a challenge to perception.
When side + side appears on the high-heeled shoes and two outside or inside conditions appear simultaneously, the robot may fail during reorientation.

\subsubsection{Experiments on reorienting through contact}
\label{Experiments on reorienting through a step}

We also tested another reorientation method, namely, contact-based reorientation, on different types of shoes (Fig. \ref{shoe_types} (a)).
The matching box heights were 110, 90, 115, 115, and 110 mm (Fig. \ref{shoe_types} (b)).
For each shoe type, the tested offsets were 5, 10, 15, 20, and 25 mm, and each case was tested 10 times. 
Additionally, the order of placement of the shoes affected the success rate of reorienting, and related experiments were conducted.

\begin{table}[htbp]
\caption{Experimental results on shoe reorienting through contact with the box at different offsets (mm).}
\label{step_results1}
\begin{center}
\resizebox{0.75\textwidth}{!}{

\resizebox{\linewidth}{!}{
\begin{tabular}{|c|c|c|c|c|c|c|}
\hline
\textbf{Offset} & \textbf{5} & \textbf{10} & \textbf{15} & \textbf{20} & \textbf{25} & \textbf{\begin{tabular}[c]{@{}c@{}}Success\\ rate\end{tabular}} \\ \hline 
Sports shoes      & 8/10  & 6/10 & 6/10 & 8/10  & 10/10 & 76\% \\
High-heeled shoes & 9/10  & 9/10 & 8/10 & 10/10 & 8/10  & 88\% \\
Leather shoes     & 10/10 & 9/10 & 6/10 & 8/10  & 9/10  & 84\% \\
% Canvas shoes      & 20\%  & 40\% & 30\% & 0\%   & 10\% \\
Sandal shoes      & 7/10  & 9/10 & 9/10 & 4/10  & 9/10  & 76\% \\ \hline
\end{tabular}
}
}
\end{center}
\end{table}

The test results for the different types of shoes at different eccentric distances are listed in Table \ref{step_results1}. 
Except for the high-heeled shoes, the success rates of the other types of shoes were not high and were unstable at different offsets. 
As for the high-heeled shoes, the side state was the most stable state, with the largest support surface and the lowest center of gravity.
Therefore, the success rate of the transformation to the side state was as high as 88\%.
For other types of shoes, the main reason for failure is that when they rotate, slide, and collide with the bottom, they bounce (such as sports and leather shoes) or over-roll (such as sandal shoes), causing them to deviate from the target position. 
When compared with sports shoes, although the total success rate of sandals is the same at 76\%, the upper sandals are softer and more elastic, and therefore, their failure forms are different. 
Overall, the success rate of the contact-based reorientation method depends on the shoe structure and material.

\begin{figure}[htbp]
  \centering
  \includegraphics[scale=0.35]{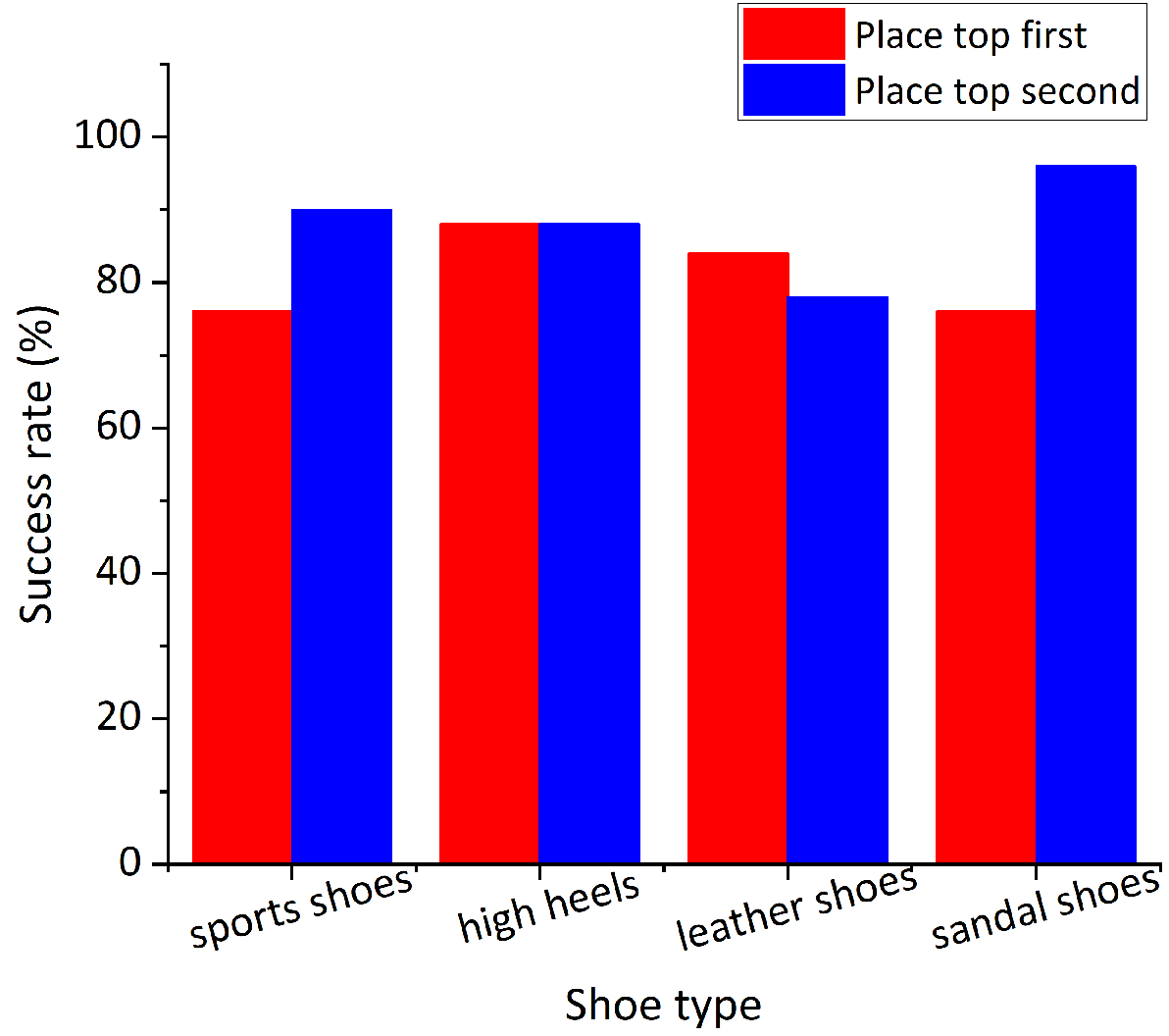}
  \caption{Comparison of success rate on the placement order for reorienting the shoe through contact.}
  \label{place_results}
\end{figure}

In order to further improve the success rate of the above reorientation method, we also tested the success rate of the reorientation method on the second-placed shoe. 
After the first shoe was placed as shown in Fig. \ref{placement_pose} (a), the top was assumed to be placed as the second shoe.
Similarly, it was placed 10 times under different offsets, and the number of successes was recorded.
As shown in Fig. \ref{place_results}, the total success rates of sports and sandal shoes significantly improve from 76\% to 90\% and 96\%, respectively.
Irrespective of the shoe type, the shoe placed first prevents the bounce or over-rotation of the second shoe. 
In this case, changing the placement order can prevent failures, especially for shoes with extremely soft or elastic uppers, such as sandals.
The success rate remained stable due to the unique structure of high heels. However, for leather shoes, changing the placement order did not improve the success rate because the reflective properties of the material seriously affected its perception.

For shoes in the top state, applying the contact-based reorientation method to the second shoe yields a higher success rate compared to applying it to the first shoe.
Therefore, we apply this contact-based reorientation method exclusively to the second shoe.
As a result, when the state combination is side + top, it can be considered a directly placeable combination, provided that the shoe in the side state is placed first, followed by the shoe in the top state.
However, perception errors caused by material reflectivity can reduce the success rate (e.g., leather shoes).

\subsection{Robotic Packing system experiments}

Based on the visual algorithm and reorientation methods described above, we conducted both qualitative and quantitative experiments on the complete packing framework, using the robotic system introduced in Section \ref{Experimental setup}.
We tested the system using four types of shoes — sports shoes, high heels, leather shoes, and sandals — which differ in size, shape, and stiffness.

\subsubsection{Shoe packing process}

Each shoe can assume one of three states: top, side, or bottom. When packing a pair of shoes, six possible state combinations must be considered: top + top, top + side, top + bottom, side + side, side + bottom, and bottom + bottom.
Among these, the side + side case must be further differentiated according to the keypoint combination defined in Equation \ref{eq7}, which results in either an inside + outside or inside + inside / outside + outside pairing.
This leads to a total of seven distinct initial combinations to consider.
For any initial combination, the packing process generally consists of two stages: a pre-placement stage and a placement stage. 
The pre-placement stage primarily utilizes primitive-based reorientation methods to transform the initial combination into a placeable state. 
Among these combinations, the side + side (inside + outside) configuration can be directly used as a placeable state without any reorientation.
Additionally, the contact-based method described earlier enables another valid placeable state: side + top. 

% \usepackage{multirow}
\begin{table}[htbp]
\caption{Toppling counts and resulting placeable states for different initial combinations with and without the contact-based reorientation method.}
\label{Ablation table}
\begin{center}
\resizebox{\textwidth}{!}{

\resizebox{2\linewidth}{!}{
\begin{tabular}{|c|c|ccccccc|}
\hline
Method                                                                              & Item                                                          & \multicolumn{1}{c|}{top + top} & \multicolumn{1}{c|}{top + side} & \multicolumn{1}{c|}{top + bottom} & \multicolumn{1}{c|}{\begin{tabular}[c]{@{}c@{}}side + side\\ (2 inside / 2 outside)\end{tabular}} & \multicolumn{1}{c|}{\begin{tabular}[c]{@{}c@{}}side + side\\ (inside + outside)\end{tabular}} & \multicolumn{1}{c|}{side + bottom} & bottom + bottom \\ \hline
\multirow{2}{*}{\begin{tabular}[c]{@{}c@{}}w/o\\ Contact-based\\ Method\end{tabular}} & \begin{tabular}[c]{@{}c@{}}Toppling\\ counts\end{tabular}     & \multicolumn{1}{c|}{2}         & \multicolumn{1}{c|}{1}          & \multicolumn{1}{c|}{2}            & \multicolumn{1}{c|}{2}                                                                            & \multicolumn{1}{c|}{0}                                                                        & \multicolumn{1}{c|}{1}             & 2               \\ \cline{2-9} 
                                                                                    & \begin{tabular}[c]{@{}c@{}}Placeable\\ state\end{tabular} & \multicolumn{7}{c|}{side + side (inside + outside)}                                                                                                                                                                                                                                                                                                             \\ \hline
\multirow{2}{*}{\begin{tabular}[c]{@{}c@{}}w/\\ Contact-based\\ Method\end{tabular}}  & \begin{tabular}[c]{@{}c@{}}Toppling\\ counts\end{tabular}     & \multicolumn{1}{c|}{1}         & \multicolumn{1}{c|}{0}          & \multicolumn{1}{c|}{1}            & \multicolumn{1}{c|}{1}                                                                            & \multicolumn{1}{c|}{0}                                                                        & \multicolumn{1}{c|}{1}             & 2               \\ \cline{2-9} 
                                                                                    & \begin{tabular}[c]{@{}c@{}}Placeable\\ state\end{tabular} & \multicolumn{4}{c|}{top + side}                                                                                                                                                                          & \multicolumn{3}{c|}{side + side (inside + outside)}                                                                                                  \\ \hline
\end{tabular}
}
}
\end{center}
\end{table}

To achieve either of these two placeable configurations, the robot must use the first reorientation method, toppling.
To better illustrate the efficiency gained from introducing the second reorientation method (contact-based), we compared the number of toppling operations required to reach a placeable state with and without its use.
Table \ref{Ablation table} presents the number of toppling steps required for each initial combination.
The introduction of the second reorientation method significantly reduced the number of toppling operations required to convert combinations such as top + top, top + side, and top + bottom into valid placeable states.
Across all seven initial combinations, the number of toppling actions decreased from 10 (without the second method) to 6 (with it).
Each initial configuration selects the shortest transformation path, and the availability of the top + side placeable state — enabled by the second reorientation method — facilitates more efficient transitions, especially for combinations involving top states.

For the two placeable configurations (side + side and top + side (inside + outside)), no toppling is needed, and the shoes can be placed directly.
Based on our evaluation of reorientation success rates (Fig. \ref{place_results}), the first shoe placed is always the one in the side state, regardless of the combination.
To ensure the correct relative orientation between the shoe and the box, the Z-axis of the the first shoe pose $P_{shoe1}^Z$ must be opposite to the Y-axis of the box pose $P_B^Y$ (Fig. \ref{placement_pose} (a)).
And it is placed at the trisection point between $P_B$ and $P_C$, closer to $P_B$.
When placing the second shoe — whether in the top or side state — it is essential that the X-direction of the second shoe $P_{shoe2}^X$ is opposite to that of the first one $P_{shoe1}^X$ (Fig. \ref{placement_pose} (b)), ensuring the correct relative orientation between the shoes.
If the second shoe is in the top state, it is placed near $P_C$ of the box with an offset of approximately 10 mm and is reoriented by exploiting contact with the box.
If the shoe is in the side state, it is placed at a similar trisection point between $P_B$ and $P_C$, but closer to $P_C$.

\begin{figure*}[htbp]
  \centering
  \includegraphics[scale=0.675]{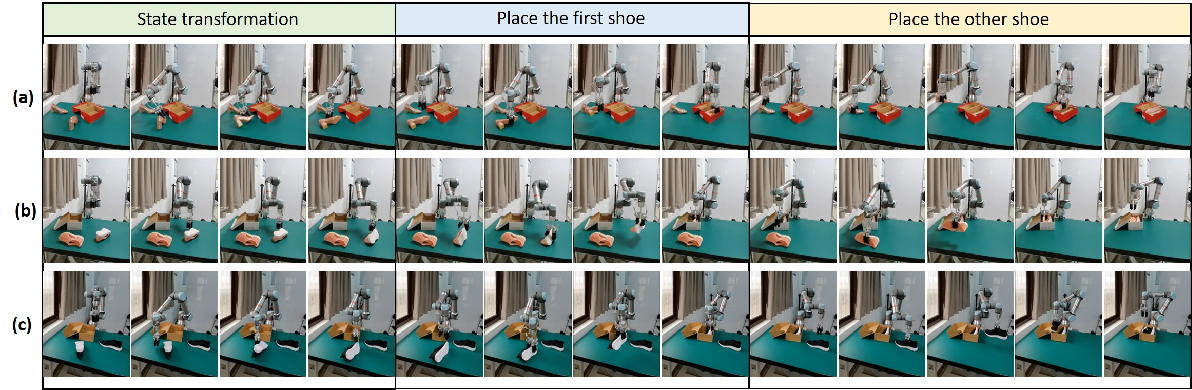}
  \caption{Packing experiments for different kinds of shoes with unknown state combinations. (a) High-heeled shoes (top + top). (b) Sandal shoes (side + bottom). (c) Sports shoes (top + bottom). A video showing examples of shoe packing can be seen at \href{https://youtu.be/SD3P4M6EOkM}{https://youtu.be/SD3P4M6EOkM}.}
  \label{qualitative}
\end{figure*}

% top+top和side+side (inside+inside)可以放在一起说。
We then focus on initial combinations that require a single toppling reorientation.
As shown in Fig. \ref{qualitative}, tested scenarios include: high-heeled shoes (top + top), Sports shoes (top + bottom), Sandals (side + bottom). 
For high heels (Fig. \ref{qualitative}(a)), when both shoes are initially in the top state, one must be toppled to change its state to side, forming a top + side combination that can be directly placed.
Since both shoes are initially in the same state, either shoe can be selected for reorientation. 
Similarly, for the side + side state combination, when the detected keypoints indicate two insides or two outsides, reorienting either shoe via a single toppling operation is sufficient to reach the desired inside + outside placeable configuration.
When one shoe is in the bottom state, such as in the top + bottom combination for sports shoes (Fig. \ref{qualitative}(c)), the robot must push the bottom-state shoe to change the combination to top + side, which can then be grasped and placed directly.
Similarly, for sandal shoes in the side + bottom state (Fig. \ref{qualitative}(b)), the robot first pushes the bottom-state shoe to transform it into the side state.
What is different is that, before pushing, the robot must determine the appropriate direction to ensure that the resulting combination is side + side (inside + outside).
For example, if the keypoint of the shoe in the side state is inside, the pushing direction should be from the outside to the inside; conversely, if the keypoint is on the outside, the pushing direction should be from the inside to the outside. 

Finally, for the bottom + bottom state — the only case requiring two toppling actions — the robot must push both shoes in opposite directions.
This transforms the configuration into a side + side (inside + outside) state that can be packed directly.

\subsubsection{Quantitative experiments}

To evaluate the performance of the robotic shoe packing system, we conducted quantitative experiments on various shoe types under different initial combinations. 
For each experimental condition, we performed ten packing trials.
In addition, owing to their distinctive geometries, leather shoes and high heels do not possess a bottom state.

\begin{table}[htbp]
\caption{Packing success rates across different shoe types and initial state combinations.}
\label{Whole packing results}
\begin{center}
\resizebox{\textwidth}{!}{

\resizebox{\linewidth}{!}{
\begin{tabular}{|cc|c|c|c|c|c|}
\hline
\multicolumn{2}{|c|}{Shoe type}                                                                                       & Sandal shoes & Sports shoes & Leather shoes & High-heeled shoes & Success rate \\ \hline
\multicolumn{1}{|c|}{\multirow{3}{*}{Group A}} & Top+side                                                             & 10/10        & 10/10        & 10/10         & 9/10              & 97.50\%      \\ \cline{2-7} 
\multicolumn{1}{|c|}{}                         & Top+top                                                              & 10/10        & 9/10         & 9/10          & 9/10              & 92.50\%      \\ \cline{2-7} 
\multicolumn{1}{|c|}{}                         & Top+bottom                                                           & 10/10        & 8/10         & N/A           & N/A               & 90.00\%      \\ \hline
\multicolumn{1}{|c|}{\multirow{4}{*}{Group B}} & \begin{tabular}[c]{@{}c@{}}Side+side\\ (inside+outside)\end{tabular} & 8/10         & 9/10         & 8/10          & 4/10              & 72.50\%      \\ \cline{2-7} 
\multicolumn{1}{|c|}{}                         & \begin{tabular}[c]{@{}c@{}}Side+side\\ (inside+inside)\end{tabular}  & 8/10         & 8/10         & 5/10          & 2/10              & 57.50\%      \\ \cline{2-7} 
\multicolumn{1}{|c|}{}                         & Side+bottom                                                          & 7/10         & 8/10         & N/A           & N/A               & 75.00\%      \\ \cline{2-7} 
\multicolumn{1}{|c|}{}                         & Bottom+bottom                                                        & 7/10         & 7/10         & N/A           & N/A               & 70.00\%      \\ \hline
\multicolumn{2}{|c|}{Success rate}                                                                                    & 85.71\%      & 84.29\%      & 80.00\%       & 60.00\%           & --           \\ \hline
\end{tabular}
}
}
\end{center}
\end{table}

The experimental results are shown in the table above. 
For ease of analysis, we categorized all initial combinations into two groups. 
Group A includes combinations that can reach the placeable state of top + side without going through the side + side configuration, namely top + side, top + top, and top + bottom. 
All other combinations, which require transitioning through the side + side configuration, are classified into Group B.
For Group A, regardless of shoe type, the top + side combination achieved the highest success rate of $97.50\%$, followed by top + top at $92.50\%$ and top + bottom at $90.00\%$. 
The high success rate of top + side is attributed to the fact that it is already in the placeable state and does not require any toppling operations, resulting in fewer steps and higher reliability. 
The limited failures observed in Group A are mainly attributed to the insufficient precision of keypoint detection, which influences both the determination of the toppling position and the grasp pose evaluation.

In Group B, the packing success rate also decreases with the number of toppling operations required for each category of shoes. 
For example, the bottom + bottom combination requires two toppling operations, making it the least successful configuration within the same shoe category. 
For instance, the success rate for sandal shoes and sports shoes in this configuration is only 7/10. 
Overall, the success rate of Group B is significantly lower than that of Group A.
The reasons for the lower success rate in Group B are both visual and physical. 
From a visual perspective, in addition to reduced keypoint detection precision, keypoint accuracy is also a major issue. 
All combinations in Group B involve a side + side state, where the high visual similarity between the medial (inside) and lateral (outside) sides of the shoes often leads to misclassification of the corresponding keypoints. 
This, in turn, results in incorrect decision-making during the planning phase.
Among all shoe types, high heels exhibit the highest side-view similarity. 
Moreover, their unique structure results in a low center of gravity in the side state, leading to extremely low toppling success rates. 
As a result, the success rate for the side+side (inside + outside) combination for high heels is only 2/10. 
In contrast, leather shoes suffer from low visual reliability due to the reflective nature of their material, which also contributes to decreased performance.

Across all initial combinations, sandal shoes achieved the highest average success rate ($85.71\%$), followed by sports shoes, leather shoes, and high heels. 
These success rates are largely influenced by the structural and material properties of each shoe type. 
For instance, the reflective surface of leather shoes impairs visual sensing, while the structure of high heels not only leads to low toppling success rates but also causes ambiguity in side-state keypoint detection. 
Consequently, high heels show the lowest overall packing success rate of only $60\%$.

In summary, the key factors affecting the packing success rate can be grouped into two categories. 
The first involves the physical properties of the shoes, including material (e.g., reflectivity) and structural characteristics, such as those of high heels, which impact both toppling performance and keypoint detection accuracy. 
The second factor is the number of required toppling operations: as this number increases, the overall success rate tends to decrease.

\section{Discusssion}
\label{Discussion}

\subsection{Discussion on perception}

We developed a keypoint-based visual perception module that can adapt to shape changes caused by differences in the type, state, and deformation. 
Furthermore, when combined with the geometric features of the objects within a class, these keypoints can be used to infer more information, such as the state, pose, and operation points, to facilitate subsequent operations.
The box is also represented by keypoints. However, a traditional visual detection method is used to obtain the keypoints because the box has a simple structure and obvious features. 
In this task, the relative pose of the shoes and the relative pose between the shoes and the box needs to be considered.

When compared to current keypoint-based manipulation methods \cite{manuelli2019kpam} \cite{gao2021kpamsc} \cite{wang2020learning} \cite{qin2020keto} \cite{xu2021affordance} \cite{robson2022keypoint}, the visual module in this study combines intraclass features to obtain more visual information. 
Information extracted in this manner exhibits unique advantages over proprietary modules (Tables \ref{state_classification}–\ref{infer_speed}) \cite{lin2022icra:centerpose} \cite{he2016deep}. 
In addition to assisting operations, the visual module also includes inspection functions such as size matching and pairing inspection. 
This is another advantage of vision as an independent module, which cannot be achieved using end-to-end models for manipulation \cite{zitkovich2023rt} \cite{ma2024survey} \cite{kim2024openvla}. 
Notably, the keypoint detection network in this visual module can be functionally replaced by other keypoint detection methods, such as large visual models \cite{huang2024rekep} \cite{fang2024keypoint}. 
However, its performance can be significantly compromised because a large model has a higher demand for computing resources and a slow update speed.

This study integrates multiple advantages of keypoint-based representation methods, including the ability to (1) adapt to large intraclass variations, (2) extract more usable information, and (3) represent the relationship between objects, which is particularly significant for multi-object operations.

The perception module in our framework is primarily based on keypoint detection. 
However, in the context of the shoe packing task, its performance is significantly influenced by the shoes' structural characteristics and materials. 
For instance, the high visual similarity between the medial (inside) and lateral (outside) sides of shoes—especially in high heels—often leads to keypoint misclassification. 
Additionally, reflective materials such as leather can interfere with visual feature extraction. 
To address these challenges, future work could incorporate depth information to enhance the detection of keypoints on shoes in the side state, and introduce adaptive lighting strategies to improve the robustness and accuracy of perception \cite{astanin2017reflective} \cite{weng2020multi} \cite{sun2023trosd}.

\subsection{Discussion on manipulation}

In the manipulation section, we proposed a two-level planner optimized for shoe packing tasks, consisting of reorientation planners and a packing task planner. 
Reorientation planners include both primitive and contact-based methods. We tested the success rates of these methods and examined the effect of the placement order. 
Based on these results, we designed a packing-task planner that efficiently transitioned a pair of shoes from any initial state to the target configuration. The experimental validation confirmed the feasibility of the proposed method.

Compared with existing shoe-packing methods \cite{perez2018automation}, \cite{gracia2017robotic} \cite{morales2014bimanual}, our approach fully considers the differences between various initial poses of the shoes and standard target poses. 
First, for multiple states of a single shoe, we defined the corresponding reorientation methods and discussed the influence of the shoe material, structure, and placement order on the success rate. 
These methods were then applied to the transition of a pair of shoes with combined states, and the shoes were placed in optimal order to efficiently complete the packing task. 
Furthermore, the soft gripper used in the experiments demonstrated excellent adaptability to irregular shapes and deformable characteristics of shoes, especially during the grasping and reorientation processes. 
This packing task provides a useful reference for packing pairs of items, particularly those with irregular shapes and varying levels of softness.

The primitive-based method is designed for various states, whereas the contact-based method targets the top states. 
In particular, for shoes in the top state, the first method achieves reorientation through interference between the gripper and the shoe surface, whereas the second method avoids manipulation of the deformable upper surface. 
Although the latter method is more efficient, its success rate is higher when applied to the second shoe (Fig. \ref{place_results}). Therefore, the combination of both methods yields the best performance. 
These reorientation methods not only address 3D deformable object reorientation, but also offer insights for multi-object manipulation. 
Notably, the use of extrinsic resources significantly enhances the success rate and efficiency of packing pairs of shoes. 
This method offers two key insights: In multi-object manipulation tasks, the contact between objects can serve as a beneficial factor for manipulation. 
Among the extrinsic resources (including contact, gravity, and inertia), a combination is sometimes more effective than using any single resource.

From the manipulation perspective, its impact on packing success is most evident during the pre-placement stage. 
On one hand, the success rate decreases as the number of toppling operations increases. 
On the other hand, for shoes with special structures—such as high heels—the success rate of toppling from the side state is extremely low. 
Among these factors, structural characteristics pose the most significant challenge. 
To tackle this, future work should consider developing specialized regrasping reorientation methods \cite{wada2022reorientbot} \cite{xu2022efficient} \cite{wan2019regrasp} \cite{wan2015reorientating} tailored for structurally complex shoes.

\section{Conclusion}
\label{Conclusion}

This study proposes a complete standard packing scheme for footwear of different sizes, shapes, and softness, including visual perception, reorientation planning, and packing task planning. 
In terms of visual perception, by adapting to large intracategory variations, a keypoint-based representation method is proposed and used to obtain further information such as the size, state, pose, and even the operation point of the shoe by combining structural characteristics of the shoe. 
Based on the visual information presented above, we used desktop and other environmental constraints to design primitive-based and contact-based reorientation methods for shoes in different states using soft grippers. 
Finally, the reorientation methods were used in the subsequent paired shoe-packing task planner to change the pairing state of the shoe to its target combination. 
We conducted qualitative and quantitative experiments to evaluate the success rate of shoe placement and to verify the feasibility of the shoe-packing task planner. 

In summary, this study further explored the advantages of keypoint representation methods and their guiding roles in operations. 
Additionally, a preliminary exploration of the reorientation method of a 3D deformable object was performed, such as using the interference between the end effector and deformable part to apply the primitive. 
Simultaneously, in multi-object manipulation tasks, the contact between objects could be used to assist manipulation including reorientation. 
Finally, this study provides a reference for packing object pairs.

As demonstrated in the experiments, the unique structure of certain shoes presents major challenges to the packing process. 
On one hand, the high visual similarity between the medial and lateral sides leads to reduced accuracy in keypoint detection. 
On the other hand, structural features significantly impact the success rate of reorientation via toppling, with high heels being the most affected in both aspects. 
To address these issues, future work could integrate depth-aware visual features to improve keypoint localization and design reorientation methods specifically adapted to structurally complex shoes.
Additionally, the packing paper prevents dust and moisture during the packing process. 
A future study will consider incorporating the placement and folding of paper into the entire packing pipeline.

% \section*{ACKNOWLEDGMENT}

% We would like to thank Poramate Manoonpong for discussions and comments on the paper.

%\appendix
%\section{Example Appendix Section}
%\label{app1}


%%       \cite{<label>} ==> [1]

%%
%Example citation, See \cite{wang2023research}.

\bibliographystyle{elsarticle-num} 
\bibliography{Reference}




% %% For numbered reference style

%   \textit{\LaTeX: a document preparation system},
%   Addison Wesley, Massachusetts,

% \end{thebibliography}

\end{document}